\documentclass[lettersize,journal]{IEEEtran}
\usepackage{amsmath,amsfonts,amssymb}
\usepackage[ruled,vlined]{algorithm2e}
\usepackage{algpseudocode}
\usepackage{array}
\usepackage{hyperref}
\usepackage{textcomp}
\usepackage{stfloats}
\usepackage{url}
\usepackage{verbatim}
\usepackage{graphicx}
\usepackage{subfigure}
\usepackage{cite}
\usepackage{xurl}
\usepackage{mathtools}
\usepackage{multirow}
\usepackage{tabularx}
\usepackage{makecell}
\usepackage{booktabs}
\usepackage{cleveref}

\crefname{equation}{}{}

\usepackage{tikz}

\begin{document}

\title{Physics-Informed Graph-Based Safe RL with Constraint-Specific Critics for Community Battery Coordination in Distribution Network} 
\author{Chenghao~Huang,
        Yue~Yang, and~Hao~Wang,~\IEEEmembership{Member,~IEEE}
\thanks{This work was supported in part by the Australian Research Council (ARC) Discovery Early Career Researcher Award (DECRA) under Grant DE230100046. (Corresponding author: Hao Wang)}
\thanks{C. Huang, Y. Yang, and H. Wang are with the Department of Data Science and AI, Faculty of IT and Monash Energy Institute, Monash University, Melbourne, VIC 3800, Australia (e-mails: \{chenghao.huang, yue.yang1, hao.wang2\}@monash.edu).}
}

\maketitle

\begin{abstract}
High penetrations of distributed energy resources (DERs) and fast-varying electric vehicle (EV) charging intensify voltage and line-limit violations in distribution networks (DNs), making safe, low-cost operation increasingly challenging.
This paper proposes a physics-informed safe deep reinforcement learning (DRL) framework for coordinating community battery energy storage systems to mitigate constraint violations while minimizing operating cost.
To represent nonlocal spatiotemporal interactions, we develop a dual-level physics-informed graph representation in which line impedance and capacity shape spatial information propagation, while power-balance, voltage-profile, and flow-reconstruction losses regularize the latent state toward control-relevant physical consistency.
Building on this representation, we develop Lagrangian Dedicated-Critic PPO (LDCPPO), which implements constraint-specific value decomposition: independent reward and cost critics preserve heterogeneous voltage- and line-loading-related learning signals and combine their advantages only during the Lagrangian policy update.
Case studies on DNs with high rooftop photovoltaic penetration and EV charging demonstrate that the proposed method reduces the key voltage-violation degree by up to 80.5\% compared with the tested baselines, while also achieving the lowest voltage-violation count and line-overload degree with competitive operating cost.
\end{abstract}


\section{Introduction}

\subsection{Background and Motivation}
Global efforts to reduce carbon emissions and achieve net-zero energy systems are driving the decentralization of electricity grids and the rapid growth of Distributed Energy Resources (DERs), especially the behind-the-meter ones such as rooftop PV and electric vehicles (EVs) \cite{IEAnetzero}.
In Australia, for example, around 40\% of households have rooftop solar, over 185,000 have battery storage \cite{pv_battery}, and EV ownership exceeded 200,000 by the end of 2024 \cite{ev}.
However, the increasing penetration of behind-the-meter DERs poses new operational challenges for distribution networks (DNs), such as voltage violations and line load capacity violations, which threaten grid safety \cite{6194986,esb_aemc}, demanding better DER coordination \cite{yang2021exploring}.
Traditional regulation devices struggle under high-DER conditions due to limited responsiveness, typically like on-load tap controllers (OLTCs) lacking flexibility, and static voltage compensators requiring escalating deployment cost.
On the other hand, behind-the-meter solutions such as inverter control \cite{8516395}, household batteries \cite{8027057}, or coupling PV with EVs \cite{7775016} offer limited and fragmented regulation capacity due to small sizes but large quantities, and intermittency, making network-wide safety difficult to comply.

To overcome these limitations, Community-scale Battery Energy Storage Systems (CBESSs) \cite{Ausgrid_CB} have emerged as shared assets and managed by Distribution System Operators (DSOs), designed to deliver grid support and community-wide benefits.
CBESSs are often installed at the neighborhood or substation level with the capacity of 200kWh-2MWh commonly \cite{DCCEEW_CB}, serving consumer communities with DERs. Compared with individual household batteries, CBESSs offer a versatile solution that can simultaneously provide voltage and power flow regulation, meet collective consumer needs, and participate in energy markets for arbitrage.
Furthermore, unlike distributed batteries, CBESSs can be owned and scheduled by DSOs, making CBESS coordination a viable solution to safe DN operation\cite{barbour2018community,kalkbrenner2019residential,mohanty2024neighborhood}. 
Australia's DSOs, known as Distribution Network Service Providers, have already piloted CBESSs for voltage support and solar hosting, such as Energy Queensland's deployment of 69 neighborhood batteries (with a total capacity of 4.41 MW/7.47 MWh) across Brisbane and southeast Queensland \cite{ARENA_CB}, demonstrating CBESSs' potential of supporting a safe, reliable, and economical energy transition under behind-the-meter DER growth.

\subsection{Literature Review and Challenges}

\begin{table*}[!t]
\centering
\footnotesize
\renewcommand\arraystretch{1.0}
\setlength{\tabcolsep}{1.8mm}
\caption{Comparison of grid-level voltage regulation methods for medium-size BESS, including business BESS and CBESS.}
\label{tab:LR}
\begin{tabular}{p{1.2cm}p{2.5cm}p{2.5cm}p{1.9cm}|p{1.7cm}p{2.0cm}p{2cm}p{1.3cm}}
\hline\hline
\multirow{2}{*}{\textbf{Ref.}}  &  \multicolumn{3}{c|}{\textbf{Problem setting (mainly about DERs)}} & \multicolumn{4}{c}{\textbf{Methodology}} \\ \cline{2-8}
 & \textbf{PV deployment (\%)}  & \textbf{PV penetration (\%)}  & \textbf{Residential nighttime EVC}  & \textbf{Solution}   & \textbf{Constraints} & \textbf{Feature correlation modeling} &  \textbf{Physics-informed} \\ \hline
 \cite{9079884}    & 27\% & 50\% & No & Control theory & Separate & No  & No \\ 
 \cite{9460799,10717491,10400788,10408412,10761093}  &  -, -, -, 40\%,18\%  & 40\%,40\%,70\%, -, -  & No & Optimization & Separate & No & No \\ \hline 
 \cite{9466427}    & 44\% & $\sim$ 50\% & No & DRL & Mixed & No & No \\ 
 \cite{8985179,su2025two}    & -, 24\% & 90\%, 30\% & No & DRL & Mixed & No & No \\ 
 \cite{11045109}    & 18\% & 70\% & No & DRL & Mixed & No & No \\ 
 \cite{9399637}    & 8\% & $\sim$ 40\%  & No & DRL & Mixed & No & No \\ 
 \cite{10113230}   & 9\% & $\sim$ 40\%  & No & DRL & Mixed & Spatial & Yes \\ 
 \cite{10923740}   & 6\%  &  15\% & No  & DRL & Partially separate & No & No \\ 
\textbf{Proposed}    & \textbf{80\%} &  \textbf{90\%} & \textbf{Yes} &  \textbf{DRL} & \textbf{Separate} &  \textbf{Spatiotemporal}  & \textbf{Yes} \\ 
\hline\hline
\end{tabular}
{\scriptsize \textit{Note:} ``-'': Not applicable, since the related works have not considered the system-wide cases, or explicitly shown the exact values. ``PV deployment ratio'': The proportion of buses in the network that host rooftop PV installation. ``PV penetration ratio'': The maximum ratio of installed PV peak capacity to the corresponding peak load at any bus in the network.}
\end{table*}

Before the emergence of DSO-owned CBESSs for network support, researchers focused on controlling BESS in power systems for economic benefits, such as arbitrage or service sale.
Early efforts primarily relied on optimization-based frameworks \cite{8781751,9460799,10400788,10717491} or heuristic methods \cite{9785634}, aimed at maximizing economic efficiency to improve energy sharing, investment returns, and flexible service participation.
But these works often do not consider physical network constraints, so more recent optimization-based studies have begun to incorporate typical network constraints, especially voltage violation, into the BESS scheduling problem \cite{10408412,10761093} as uncoordinated operation of DERs can harm grid safety.
Moreover, \cite{9079884} proposes a two-stage coordinated control strategy that centrally manages OLTC adjustments while locally controlling multiple BESS units for fast active power response, aiming to minimize voltage deviations.
Nevertheless, optimization-based approaches depend on accurate system models, making them less suitable for highly dynamic and uncertain environments, which however are increasingly common in distribution networks with volatile DERs.

As energy systems become more decentralized and uncertain, DRL offers a learning-based approach that can learn causal control policies from historical interactions and respond rapidly to changing operating conditions during execution.
In the context of BESS scheduling, DRL has demonstrated strong potential in handling stochasticity in renewable output, user demand, and electricity prices \cite{10513300,9493724,li2023deep,li2024temporal,10680378} via exploring flexible strategies and optimizing long-term objectives under uncertain conditions.
Furthermore, some DRL-based BESS controlling methods account for the physical constraints of DNs.
Based on DRL's reward-driven nature, some works integrated physical constraints into the reward function for long-term optimization \cite{8985179,9399637,9466427,10113230}, in which the work of \cite{10113230} integrates structured graph information from the DN physics into the DRL training process to enhance decision robustness towards voltage violation.
However, merely embedding physical constraints into the reward signal remains insufficient, as it can lead to voltage violations under highly dynamic or extreme conditions.

Recent RL studies have also explored multi-critic designs. 
However, only a limited number of studies have adopted such multi-critic safe RL structures for power system operation.
The work of \cite{10923740} developed a hybrid day-ahead and real-time framework combining optimization-based prediction and the constrained DRL paradigm. The study in \cite{kim2026safe} used multiple critics for heterogeneous operational objectives together with iterative safe exploration. 
However, for DN operation with heterogeneous voltage- and line-loading constraints, these methods do not explicitly isolate the long-term learning signals of individual physical constraint costs within a unified Lagrangian policy-optimization framework.

A brief overview of existing CBESS scheduling methods is summarized in Table \ref{tab:LR}, illustrating that CBESS coordination for network constraint maintenance under rich DERs remains an important gap in current research.
Compared with prior works, our study considers a high-PV deployment scenario (80\% bus coverage) with elevated PV penetration (up to 90\%) and explicitly incorporates residential nighttime EV charging demand.
Such settings reflect emerging feeders in regions with aggressive rooftop PV adoption and rapid EV uptake, where distributed PV systems are densely installed across residential buses.
High PV deployment significantly increases spatial voltage coupling, making local voltage deviations propagate feeder-wide.
High PV penetration intensifies reverse power flow and overvoltage risk, shrinking the voltage-feasible action region.
Introducing residential nighttime EV charging creates conflicting operating regimes: daytime overvoltage driven by PV surplus and nighttime undervoltage caused by concentrated charging demand.

Based on the literature review, we identify three fundamental challenges:
\begin{itemize}
\item \textit{Mixed constraint feedback under multi-CBESS constraint-aware learning:}  
When multiple CBESS units operate, their charging and discharging actions interact through network voltage and line loading dynamics, causing constraint violations at different buses and lines to occur concurrently and interact.  
In many DRL studies, different constraint signals are aggregated via reward shaping or weighted penalties into a single feedback signal. This obscures which constraint is active and how severe individual violations are, thus weakening constraint awareness and destabilizing training. 
A key challenge is therefore to design a constraint-aware learning framework that separates reward from constraints and further decomposes constraints for independent estimation.

\item \textit{Insufficient physical awareness in purely data-driven policies:}  
Although CBESS can regulate power injections to support grid operation, their large capacity means inappropriate dispatch may induce voltage fluctuations or line stress.  
Purely data-driven policies may not internalize power-flow physics and electrical couplings, leading to unstable or oscillatory training behavior. Incorporating physics-informed structural priors into representation learning is therefore important for improving training robustness and convergence consistency.

\item \textit{Nonlocal spatiotemporal interactions under large-scale CBESS dispatch:}  
Due to electrical coupling in distribution networks, CBESS actions at one location can affect voltages and line loadings across multiple buses. These effects also accumulate over time through network dynamics and battery state evolution.  
Effective coordination thus requires models that capture both spatial dependencies and temporal dynamics. Policies that ignore such spatiotemporal structure may lead to suboptimal or unstable multi-CBESS coordination.

\end{itemize}

The resulting methodological question is how to construct a control state that retains nonlocal power-flow relationships while separately assigning long-term credit to heterogeneous operational constraints. These two difficulties motivate the dual-level PISTG representation and LDCPPO value decomposition developed in this work.

\subsection{Main Work and Contributions}

We study the coordinated operation of CBESSs in a DN with rich rooftop PV and residential nighttime EV charging, targeting reduced network-wide physical constraint violations and lower DN-level operating cost under high adoption of behind-the-meter DERs.
To meet these requirements, we propose a DRL framework that (1) embeds power-flow priors into grid-state representation through a physics-informed spatiotemporal graph encoder, to anticipate spatially coupled dynamics for supporting smarter decision-making;
(2) casts CBESS control as a Constrained Markov Decision Process (CMDP) with multiple operational-constraint costs and optimizes economic return subject to expected constraint-cost budgets; (3) implements constraint-specific value decomposition within a Lagrangian policy-optimization framework.
The main contributions are summarized as follows.
\begin{itemize}

\item \textbf{Problem-level novelty in learning-based safe multi-CBESS coordination.}
We investigate a DSO-side learning problem for coordinating multiple CBESSs under jointly intensified rooftop-PV and residential nighttime-EV risks. The problem requires learning a causal coordination policy when daytime PV surplus and nighttime EV charging create alternating overvoltage, undervoltage, and line-loading risks, while multiple CBESS actions are nonlocally coupled through power flows and temporally coupled through PV generation, EV demand, electricity prices, and battery SOC dynamics. This setting raises the distinct challenge of achieving rapid policy execution while retaining awareness of heterogeneous voltage- and line-loading-related constraints.

\item \textbf{Dual-level physics-informed spatiotemporal representation.}
We develop a Physics-Informed Spatiotemporal Graph (PISTG) encoder that incorporates distribution-network physics at two complementary levels. At the graph-structure level, line resistance, reactance, and capacity shape how information propagates across electrically coupled buses; at the latent-representation level, consistency in the Kirchhoff's Current Law (KCL), voltage-drop consistency, and voltage/branch-flow reconstruction regularize what control-relevant physical relationships are preserved. This design enables the policy to represent both the propagation of nonlocal network effects and their physical consequences for coordinated CBESS control.

\item \textbf{Constraint-specific value decomposition for safe policy optimization.}
We develop LDCPPO to preserve the distinct long-term learning signals of the economic objective and individual operational constraints. Separate critics estimate the economic return, voltage-violation degree, voltage-violation count, and line-overload degree using independent return targets, and their advantages are combined only during the Lagrangian actor update. This design avoids heuristic reward aggregation and reduces interference and credit-assignment ambiguity among heterogeneous constraint signals.

\item \textbf{Effective constraint--economy performance with real-time policy execution.}
Extensive simulations on the IEEE 33- and 69-bus systems demonstrate the effectiveness and scalability of the proposed framework. Compared with all tested baselines, PISTG-LDCPPO reduces the voltage violation by approximately $50\%$, and achieves the lowest voltage-violation count and line-overload degree, while maintaining competitive operating cost and real-time responsiveness. These results demonstrate that the physics-informed representation and constraint-specific value learning provide complementary benefits, supporting their problem-driven coupling.
\end{itemize}

The remainder of this paper is organized as follows. 
Section \ref{Sec:SM} models the CBESS coordination problem.
Section \ref{Sec:M} introduces the developed method. 
Section \ref{Sec:E} provides experimental results and analysis to validate the effectiveness of our method.
Section \ref{Sec:C} draws the conclusion.

\section{System Modeling}\label{Sec:SM}
\begin{figure}[t]
\centering
\includegraphics[width=0.48\textwidth]{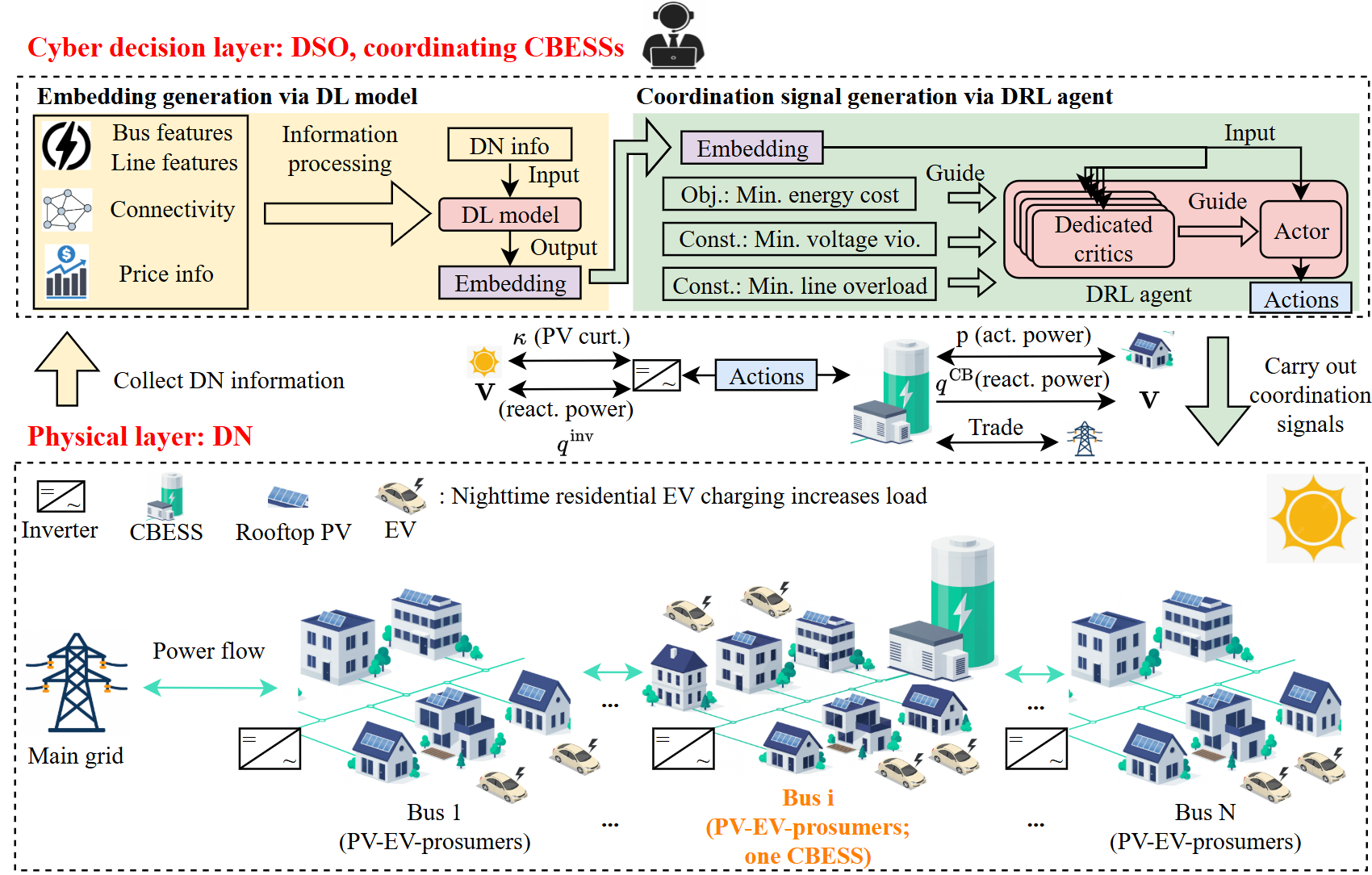}
\caption{System model of CBESS coordination via a DSO's centralized training in a rich-DER DN.} 
\label{fig:system model}
\end{figure}

We consider a radial DN with highly integrated rooftop PV and residential nighttime EV charging across multiple buses, where several CBESS units are strategically deployed at nodes within DER-dense regions. In such settings, voltage-limit compliance and energy efficiency can be challenged by the temporal and spatial mismatch between daytime PV generation and nighttime EV charging demand, while CBESS operation may also affect line loading across the network. An overview of the modeled system is illustrated in Fig.~\ref{fig:system model}.

\subsection{System Description}

We consider a radial DN $(\mathcal{N},\mathcal{L})$ managed by a DSO over intra-day periods $t \in \mathcal{T}=\{1,\dots,T\}$.
Then, we successively define CBESS attributes, high DER-penetration buses, bus-level inverters, and DN-level principles.

\subsubsection{CBESS}

To coordinate the rooftop PV shift for meeting both system safety and energy efficiency, we consider a set of CBESSs owned by the DSO, denoted by $\mathcal{M}$. 
Each CBESS $m\in\mathcal{M}$ is connected at bus $\xi(m)$ and characterized by charge/discharge efficiencies $(\eta^{\text{ch}}_m,\eta^{\text{dis}}_m)$, charge/discharge limits $(\overline{P}^{\text{ch}}_m,\overline{P}^{\text{dis}}_m)$, and state-of-charge (SOC) bounds $[\underline{SOC}_m,\overline{SOC}_m]$.
The charging/discharging power variables are $p^{\text{ch}}_{m,t}$ and $p^{\text{dis}}_{m,t}$, the active power of CBESS is $p^{\text{CB}}_{m,t} = p^{\text{dis}}_{m,t}-p^{\text{ch}}_{m,t}$, the reactive support is $q^{\text{CB}}_{m,t}$, and the stored energy is $E_{m,t}$ with capacity $E_m^{\text{cap}}$.
Their dynamics and constraints are given by:
\begin{subequations}\label{eq: battery}
\begin{align}
    &E_{m,t+1} = E_{m,t} + \eta^{\text{ch}}_m p^{\text{ch}}_{m,t}\Delta t - \frac{1}{\eta^{\text{dis}}_m}p^{\text{dis}}_{m,t}\Delta t, \label{eq: cbess dynamics}\\
    &SOC_{m,t} = \frac{E_{m,t}}{E_m^{\text{cap}}}, \quad \underline{SOC}_m \leq SOC_{m,t} \leq \overline{SOC}_m, \label{eq: soc}\\
    &0 \leq p^{\text{ch}}_{m,t} \leq \overline{P}^{\text{ch}}_m, \quad
    0 \leq p^{\text{dis}}_{m,t} \leq \overline{P}^{\text{dis}}_m, \label{eq: cd rate cap}\\
    &p^{\text{ch}}_{m,t}\cdot p^{\text{dis}}_{m,t} = 0, \label{eq: cd exclusivity}\\
    &(p^{\text{dis}}_{m,t}-p^{\text{ch}}_{m,t})^2 + (q^{\text{CB}}_{m,t})^2 \leq (S^{\text{CB}}_m)^2, \label{eq: cbess thermal limits}\\
    &E_{m,0} = E^{\text{init}}_m, \label{eq: cbess init}
\end{align}
\end{subequations}
in which Eq. \eqref{eq: cbess dynamics} defines CBESS dynamics.
Eq. \eqref{eq: soc} defines SOC boundaries.
Eq. \eqref{eq: cd rate cap} defines charging/discharging rate boundaries.
Eq. \eqref{eq: cd exclusivity} defines charging/discharging exclusivity.
Eq. \eqref{eq: cbess thermal limits} defines the CBESS loading limit, where $S^{\text{CB}}_m$ is inverter apparent power rating.
Eq. \eqref{eq: cbess init} defines the CBESS storage initialization at each day's beginning, where $E^{\text{init}}_m$ is the initial volume of the $m$-th CBESS and generally set as 50\% of $\overline{SOC}_m$.

\subsubsection{High DER-Penetration Buses}

Let $\mathcal{N}^{\text{DER}}\subseteq\mathcal{N}$ denote the set of high-DER-penetration buses, characterized by both significant rooftop PV installation and residential EV charging demand. 
This reflects regions with high adoption of DERs.
To avoid overly dispersed DER deployment that weakens spatial interaction effects, high-DER buses are primarily located around CBESS-connected buses, forming locally intensive DER regions.

For each $i\in\mathcal{N}^{\text{DER}}$, we define time-varying active and reactive load profiles $p^{\text{load}}_{i,t}$, $q^{\text{load}}_{i,t}$, and available PV generation profile $\tilde p^{\text{PV}}_{i,t}$.
For buses outside $\mathcal{N}^{\text{DER}}$, standard load profiles are assumed without PV generation and EV charging.

\subsubsection{PV Inverters}

The PV system at each bus $i\in\mathcal{N}^{\text{DER}}$ is assumed to have an inverter that regulates rooftop PV output and provides reactive support.
The PV curtailment ratio is denoted by $\kappa_{i,t}\in[0,1]$, where $\kappa_{i,t}=0$ means no curtailment and $\kappa_{i,t}=1$ means full curtailment. The curtailed PV output is:
\begin{align}\label{eq:pv curt}
    p^{\text{PV}}_{i,t} = (1-\kappa_{i,t})\,\tilde p^{\text{PV}}_{i,t}.
\end{align}
The inverter can also inject or absorb reactive power $q^{\text{inv}}_{i,t}$, bounded by its apparent power rating $S^{\max}_i$:
\begin{align}\label{eq:inverter}
    (p^{\text{PV}}_{i,t})^2 + (q^{\text{inv}}_{i,t})^2 
    \le (S^{\max}_i)^2.
\end{align}
where $\tilde p^{\text{PV}}_{i,t}\ge 0$ denote the original PV output at bus $i$.

\subsubsection{DN}

We model the radial DN $(\mathcal{N},\mathcal{L})$ using Baran-Wu DistFlow formulation~\cite{25627}.
The DN $(\mathcal{N},\mathcal{L})$ can be equivalently described by an adjacency matrix $\mathbf{A}\in \{0,1\}^{|\mathcal{N}|\times|\mathcal{N}|}$, where each entry is defined as:
\begin{align}\label{eq:adjacency}
    A_{ij} = \begin{cases} 1, \quad \text{if } (i,j) \in \mathcal{L},\\ 0, \quad \text{otherwise}.
    \end{cases}
\end{align}
Since the network is radial, $\mathbf{A}$ is sparse and symmetric, with exactly one unique path between any two buses. Each line $(i,j)\in\mathcal{L}$ denotes a branch from parent bus $i$ to child bus $j$. The corresponding branch flows are denoted by $(p_{ij,t},q_{ij,t})$. 
Bus $0$ denotes the substation with fixed voltage magnitude $V_{0,t}=V_0$.

For buses without CBESS, the balance reduces to the standard form between demand, PV, and power flow. 
For buses hosting CBESS, additional charge/discharge terms are included:
\begin{subequations}\label{eq: bus power}
\begin{align}
    &\sum_{j:(j,i)\in\mathcal{L}} p_{ji,t} + p^{\text{PV}}_{i,t} + \sum_{m:\xi(m)=i} p^{\text{CB}}_{m,t} = p^{\text{load}}_{i,t} + \sum_{j:(i,j)\in\mathcal{L}} p_{ij,t}, \label{eq: a bus power balance}\\
    &\sum_{j:(j,i)\in\mathcal{L}} q_{ji,t} + q^{\text{inv}}_{i,t} + \sum_{m:\xi(m)=i} q^{\text{CB}}_{m,t}
    = q^{\text{load}}_{i,t} + \sum_{j:(i,j)\in\mathcal{L}} q_{ij,t}, \label{eq: r bus power balance}
\end{align}
\end{subequations}
where $(j,i)$ denotes all downstream branches of the $j$-th bus, and $\xi(m)$ denotes the bus at which the $m$-th CBESS is connected.

Based on the classical Baran-Wu DistFlow model, the branch power balance and voltage magnitude relations along line $(i,j)$ are described as follows:
\begin{subequations}\label{eq: distflow const}
\begin{align}
    &p_{ij,t} = \sum_{k:(j,k)}p_{jk,t} + p_{j,t} + r_{ij}\ell_{ij,t},\\
    &q_{ij,t} = \sum_{k:(j,k)}q_{jk,t} + q_{j,t} + x_{ij}\ell_{ij,t},\\
    &V^2_{j,t} = V^2_{i,t} - 2(r_{ij}p_{ij,t}+x_{ij}q_{ij,t}) + (r^2_{ij}+x^2_{ij})\ell_{ij,t}, \label{eq: voltage droop} \\
    &\ell_{ij,t} V^2_{i,t} = p^2_{ij,t} + q^2_{ij,t},\\
    &  \underline{V} \leq V_{i,t} \leq \overline{V}, \quad \forall i\in\mathcal{N}. \label{eq: voltage limit}\\
    & p_{ij,t}^2 + q_{ij,t}^2 \leq \overline{S}_{ij}^2, \quad \forall (i,j)\in\mathcal{L}, \label{eq: branch thermal limit}
\end{align}
\end{subequations}
where $r_{ij}$ and $x_{ij}$ are the resistance and reactance of the line.
In principle, bus voltages are required to remain within admissible bounds as shown in Eq.~\eqref{eq: voltage limit}, and loads on branches are required to remain within line load limits as shown in Eq.~\eqref{eq: branch thermal limit}, where $\overline{S}_{ij}$ is the apparent capacity limit.

\subsection{Objective}

This problem aims to learn a coordination strategy for multiple CBESS units operated by the DSO under high penetration of behind-the-meter DERs in a radial DN.
To clarify the target problem before introducing the CMDP formulation, the task can be interpreted as a finite-horizon stochastic coordination problem. Let $\omega$ denote the uncertain operating scenario, including rooftop PV availability, active/reactive load demand, residential EV charging demand, wholesale electricity price, and the train--test distribution mismatch caused by sampled daily profiles. The DSO aims to learn a causal policy $\pi$ that minimizes the expected daily economic cost while limiting network constraint violations:
\begin{subequations}\label{eq: problem formulation}
\begin{align}
    \min_{\pi}\quad
    & \mathbb{E}_{\omega}\left[
    \sum_{t\in\mathcal{T}}
    \left(
    f^{\mathrm{BD}}(t)+f^{\mathrm{EC}}(t)
    \right)
    \right], \label{eq: objective}\\
    \mathrm{s.t.}\quad & \eqref{eq: battery},~\eqref{eq:pv curt},~\eqref{eq:inverter},~\eqref{eq: bus power},~\eqref{eq: distflow const},\\
    & \mathbb{E}_{\omega}\left[\sum_{t\in\mathcal{T}} f^{\mathrm{VD}}(t)\right]\leq d_{\mathrm{VD}},\\
    & \mathbb{E}_{\omega}\left[\sum_{t\in\mathcal{T}} f^{\mathrm{VN}}(t)\right]\leq d_{\mathrm{VN}},\\
    & \mathbb{E}_{\omega}\left[\sum_{t\in\mathcal{T}} f^{\mathrm{LO}}(t)\right]\leq d_{\mathrm{LO}}.
\end{align}
\end{subequations}

Here, $f^{\mathrm{EC}}(t)$ and $f^{\mathrm{BD}}(t)$ denote the energy cost of the whole DN and the CBESS degradation cost, respectively, while $f^{\mathrm{VD}}(t)$, $f^{\mathrm{VN}}(t)$, and $f^{\mathrm{LO}}(t)$ denote the voltage-violation degree, voltage-violation count, and line-overload degree, whose expected cumulative tolerances are specified by $d_{\mathrm{VD}}$, $d_{\mathrm{VN}}$, and $d_{\mathrm{LO}}$, respectively. These terms correspond to the reward and operational-constraint-cost definitions in the subsequent CMDP formulation. These constraints are evaluated at the network level through power-flow relationships, and therefore are coupled across spatially distributed buses. The additional challenge arises from the complex spatiotemporal interactions under large-scale DER integration and multi-CBESS dispatch, which makes constraint evaluation and economic optimization tightly interdependent. Therefore, the above optimization-oriented form is used to clarify the physical and economic target, while the actual sequential decision-making problem is formulated as a constrained Markov decision process (CMDP) in Section~\ref{sec:cmdp}.

\section{Methodology}\label{Sec:M}

Building on Section~\ref{Sec:SM}, this section presents the proposed framework. First, the dual-level Physics-Informed Spatiotemporal Graph (PISTG) representation integrates line-parameter-weighted graph propagation with physics-regularized latent representation learning. Second, the CBESS coordination task is formulated as a CMDP, where PISTG embeddings define the states, CBESS/inverter set-points define the actions, and voltage and line-loading violations are modeled as separate operational-constraint costs. Third, Lagrangian Dedicated-Critic PPO (LDCPPO) implements constraint-specific value decomposition using independent reward and cost critics with adaptive Lagrangian multipliers to optimize economic return subject to expected constraint-cost budgets. Together, these components support physically informed state representation and structured multi-constraint policy optimization.

\begin{figure}[t]
\centering
\includegraphics[width=0.49\textwidth]{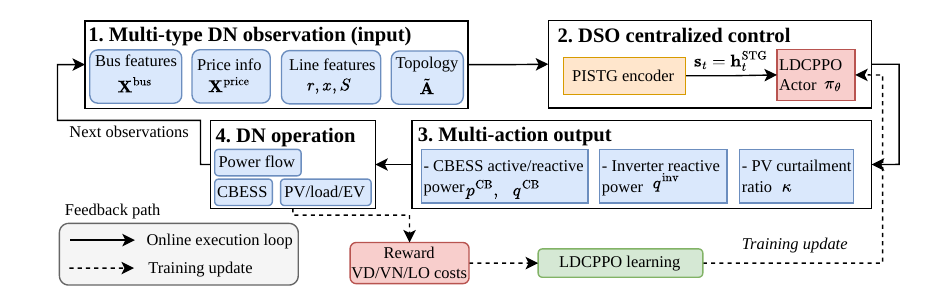}
\caption{Problem-driven coupling of the dual-level PISTG representation (Stage 1) and constraint-specific LDCPPO policy optimization (Stage 2) for DSO-coordinated multi-CBESS control.}
\label{fig:methodology brief}
\end{figure}

Figure~\ref{fig:methodology brief} first provides an abstracted overview of the overall framework.
The DSO receives multi-type DN observations, encodes them through the PISTG state encoder, generates coordinated CBESS/inverter/PV-curtailment actions through the LDCPPO actor, and applies these actions to the distribution-network environment. The resulting next observations close the online execution loop, while the reward and operational-constraint costs are fed to the LDCPPO learning process for training-time policy update only.

\begin{figure*}[t]
\centering
\includegraphics[width=0.99\textwidth]{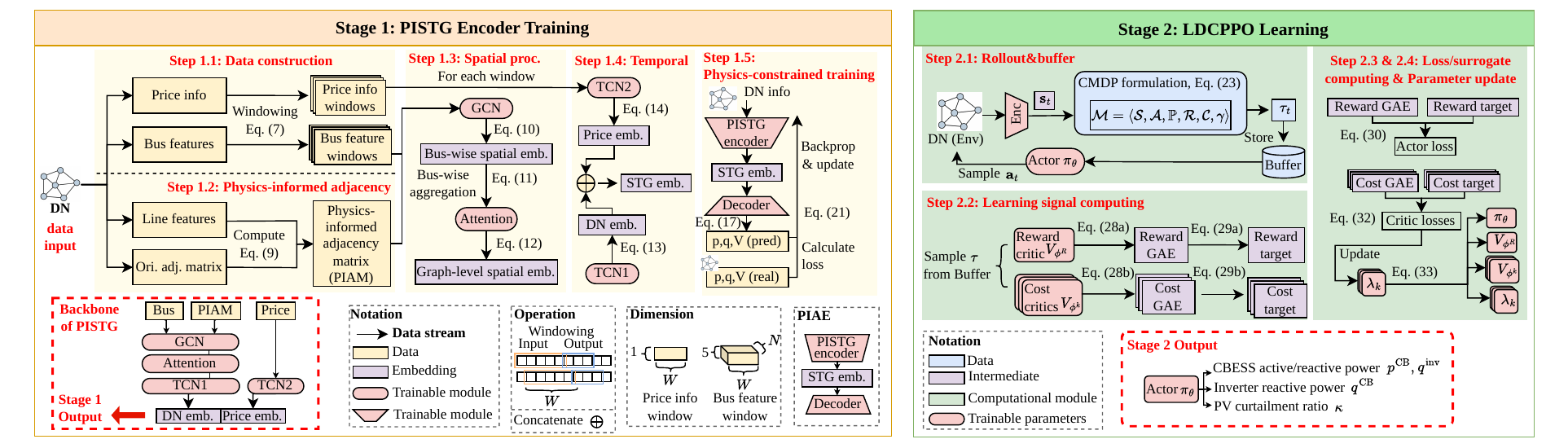}
\caption{Problem-driven coupling of the dual-level PISTG representation (Stage 1) and constraint-specific LDCPPO policy optimization (Stage 2) for DSO-coordinated multi-CBESS control.}
\label{fig:methodology}
\end{figure*}

The overall workflow is illustrated in Fig.~\ref{fig:methodology}. 
The main body of the figure further separates the framework into Stage~1, i.e., physics-regularized PISTG representation learning, and Stage~2, i.e., LDCPPO learning with rollout, learning-signal computation, and actor/critic/multiplier updates. The PISTG backbone is also highlighted to show how bus features, the physics-informed adjacency matrix, and price information are processed by GCN, attention, and TCN modules.

\subsection{Dual-Level Physics-Informed Spatiotemporal Representation}\label{sec:graph}

To capture DN spatiotemporal dependencies, we design a dual-level physics-informed spatiotemporal graph (PISTG) representation that incorporates electrical coupling into graph propagation and physical consistency into latent representation learning.
Briefly, the encoder builds a physics-informed adjacency matrix (PIAM) reflecting line coupling, extracts spatial features from power-flow patterns and temporal features via temporal convolutions, and is trained under an auto-encoder paradigm with physics-informed losses.
PIAM and the physics-informed losses play complementary roles: PIAM shapes the electrical message-passing structure, while the KCL- and voltage-drop-related losses regularize the physical consistency of the learned latent representation.

\subsubsection{Data Structure Construction}
First, a spatiotemporal graph snapshot is constructed by windowed bus features and prices to characterize the DN dynamics.
For the temporal dimension, we construct a window of length $W$ (e.g., $W$ consecutive 5-minute intervals), which stacks recent observations.
Further on the spatial side, we organize the feature windows of all the buses, each of which contains active/reactive load, PV generation, voltage, and CBESS energy (set to zero if no CBESS is connected to the $i$-th bus), as a three-dimensional tensor.
The bus features $\mathbf{X}^{\text{bus}}_{t-W+1:t}$ and temporal price window $\mathbf{X}^{\text{price}}_{t-W+1:t}$ are formulated as follows:
\begin{subequations}\label{eq:data construction}
\begin{align}
    &\mathbf{X}^{\text{bus}}_{t-W+1:t}=\Big[\big[\mathbf{x}^{\text{bus}}_{i,t'}\big]_{t'=t-W+1}^t \Big]_{i\in\mathcal{N}} \in \mathbb{R}^{W\times |\mathcal{N}|\times d_{\text{bus}}},\\
    &\mathbf{x}^{\text{bus}}_{i,t'} = \Big[p^{\text{load}}_{i,t'},q^{\text{load}}_{i,t'},p^{\text{PV}}_{i,t'},V_{i,t'},\underbrace{E_{i,t'}}_{0, \text{if }\nexists m:\xi(m)=i}\Big]^{\top} \in \mathbb{R}^{d_{\text{bus}}},\\
    &\mathbf{X}^{\text{price}}_{t-W+1:t} = [\varphi^{\text{wh}}_{t'}]_{t'=t-W+1}^t \in \mathbb{R}^{W \times 1},
\end{align}    
\end{subequations}
where $d_{\text{bus}}$ is the number of the bus-level features at one moment. 
The bus-level variables, including load, PV generation, voltage magnitude, and CBESS energy state, are assumed to be obtained from pseudo-measurements, while the feeder topology and line parameters are assumed to be available from the DSO's network model.

Additionally, we augment edges with line features:
\begin{subequations}
\begin{align}
    &\mathbf{B} = \big[\mathbf{b}_{ij}\big]_{(i,j)\in\mathcal{L}} \in \mathbb{R}^{|\mathcal{L}|\times d_{\text{line}}},\\
    &\mathbf{b}_{ij} = [r_{ij}, x_{ij}, \overline{S}_{ij}]^{\top} \in \mathbb{R}^{d_{\text{line}}},
\end{align}    
\end{subequations}
where $d_{\text{line}}$ is the number of the line-level features.

\subsubsection{PI Adjacency Matrix}
We encode the line attributes in $\mathbf{B}$ into the edge weights of $\mathbf{A}$, thereby constructing a physics-informed (PI) adjacency matrix $\tilde{\mathbf{A}}$.
Compared with a binary graph, $\tilde{\mathbf{A}}$ preserves topology while capturing the relative electrical coupling strength across connected buses, thus providing a more informative structural prior for physically consistent CBESS coordination.
The PI adjacency matrix $\tilde{\mathbf{A}}$ is devised based on Eq. \eqref{eq:adjacency} as:
\begin{subequations}
\begin{align}\label{eq:pi adjacency}
    &\tilde{\mathbf{A}} = [\tilde{A}_{ij}]_{i,j \in \mathcal{N}} \in \mathbb{R}^{|\mathcal{N}|\times|\mathcal{N}|},\\
    &\tilde{A}_{ij} = \begin{cases} \sigma [\psi(\mathbf{b}_{ij})], \quad \text{if } (i,j) \in \mathcal{L},\\ 0, \quad \text{otherwise},
    \end{cases}\\
    &\psi(\mathbf{b}_{ij}) = \underbrace{-\log (\varepsilon + \sqrt{r^2_{ij}+x^2_{ij}})}_{\text{Impedance}\downarrow\text{,ECS}\uparrow} - \underbrace{\frac{\nu}{\overline{S}_{ij}}}_{\text{Capacity}\uparrow\text{,ECS}\uparrow}.
\end{align}
\end{subequations}
Here, $\psi(\mathbf{b}_{ij})$ is a physically derived coupling score that measures the relative electrical strength of line $(i,j)$ based on its impedance and capacity. The sigmoid function $\sigma(\cdot)$ then maps this score smoothly into a bounded edge weight, preserving the relative coupling strength while maintaining numerical stability.
In \eqref{eq:pi adjacency}, $\varepsilon>0$ is a small constant used to avoid numerical instability when the electrical distance $\sqrt{r_{ij}^{2}+x_{ij}^{2}}$ approaches zero. The coefficient $\nu>0$ controls the contribution of the capacity term $\frac{1}{\overline{S}_{ij}}$ in the coupling score. 
As a result, $\psi(\mathbf{b}_{ij})$ increases when a line has lower impedance and/or larger capacity, leading to a larger PI edge weight in $\tilde{\mathbf{A}}$.
The encoder therefore takes $(\mathbf{X}^{\text{bus}}_{t-W+1}, \mathbf{X}^{\text{price}}_{t-W+1}, \tilde{\mathbf{A}})$ as inputs in a single forward pass.

\subsubsection{Spatial Processing}
We use GCN since it naturally propagates bus-level signals along electrical couplings encoded in $\tilde{\mathbf{A}}$.
Given bus-level features $\mathbf{X}_{t'} \in \mathbb{R}^{|\mathcal{N}|\times d_{\text{bus}}}$ at time $t'$, the spatial embedding $\mathbf{H}^{(l)}_{t'}$ is computed by GCN layers $\omega_{\text{GCN}}$:
\begin{subequations}\label{eq:gcn}
\begin{align}
    &\mathbf{H}^{(l+1)}_{t'} = \sigma\Big(\tilde{\mathbf{D}}^{-\frac{1}{2}} \tilde{\mathbf{A}} \tilde{\mathbf{D}}^{-\frac{1}{2}} \mathbf{H}^{(l)}_{t'} \omega_{\text{GCN}}^{(l)}\Big), \, \mathbf{H}^{(0)}_{t'} = \mathbf{X}_{t'},\\
    &\tilde{\mathbf{D}} = [\tilde{D}_{ii}]_{i\in\mathcal{N}}, \quad \tilde{D}_{ii} = \sum_{j\in \mathcal{N}} \tilde{A}_{ij}, \label{eq:diagonal matrix}
\end{align}    
\end{subequations}
where $\tilde{\mathbf{A}}$ is the PI adjacency matrix, and $\tilde{\mathbf{D}}$ is the corresponding diagonal degree matrix, whose entries $\tilde{D}_{ii}$ are computed by Eq.~\eqref{eq:diagonal matrix} as the sum of PI edge weights connected to bus $i$.
The index $l$ denotes the layer number of $\omega_{\text{GCN}}$, $L$ is the total depth, and $\omega^{(l)}_{\text{GCN}}$ is the learnable weight matrix at layer $l$. 
This formulation propagates information through physically weighted neighbors and applies nonlinear activation $\sigma(\cdot)$.
Its effectiveness stems from smoothing and mixing bus-level signals along real electrical pathways, thus embedding DN structure directly into bus-level embedding.

After $L$ layers, the bus-wise spatial embedding $\mathbf{H}^{(L)}$ is obtained.
To produce a graph-level summary where more critical buses will be more influential, we apply attention-based pooling on bus-wise embeddings:
\begin{subequations}\label{eq:att}
\begin{align}
    &\mathbf{h}^{\text{S}}_{t'} = \sum_{i\in\mathcal{N}} \alpha_{i,t'} \mathbf{h}^{(L)}_{i,t'} \in \mathbb{R}^{d_{\text{S}}},\\
    &\alpha_{i,t'} = \frac{\exp (\omega_{\text{att}}\mathbf{h}^{(L)}_{i,t'})}{\sum_j \exp (\omega_{\text{att}} \mathbf{h}^{(L)}_{j,t'})},
\end{align}
\end{subequations}
where $\omega_{\text{att}}$ is a trainable vector for attention-based weighting, deciding which buses are more critical for the final summary (e.g., those near violations). 
As a result, the encoder focuses on the most influential buses when forming $\mathbf{h}^{\text{S}}_{t'} \in \mathbb{R}^{d_{\text{S}}}$.
Finally, we get the graph-level summarized spatial embedding $\mathbf{H}^{\text{S}}_t$ via collecting over the window:
\begin{align}
    \mathbf{H}^{\text{S}}_t = \big[\mathbf{h}^{\text{S}}_{t-W+1};...;\mathbf{h}^{\text{S}}_t\big] \in \mathbb{R}^{W \times d_{\text{S}}},
\end{align}
which compresses the system into a sequence of length $W$ with embedding dimension $d_{\text{S}}$.

\subsubsection{Temporal Processing}

To capture temporal dependencies such as fast PV/load fluctuations, TCN, whose causal and dilated convolutions can extract both short- and long-term correlations in parallel while avoiding recurrence, is incorporated to process the sequence of graph-level vectors into spatiotemporal embedding: 
\begin{align}\label{eq:tcn1}
    \mathbf{h}^{\text{DN}}_{t} = \omega^{\text{DN}}_{\text{TCN}}( \mathbf{H}^{\text{S}}_t) \in \mathbb{R}^{d^{\text{DN}}_{\text{ST}}}.
\end{align}
In addition, the price sequence $\mathbf{X}^{\text{price}}_{t-W+1:t}$ is processed through a separate TCN encoder to capture its temporal regularities:
\begin{align}\label{eq:tcn2}
    \mathbf{h}^{\text{price}}_{t} = \omega^{\text{price}}_{\text{TCN}}\Big(\mathbf{X}^{\text{price}}_{t-W+1:t}\Big) \in \mathbb{R}^{d^{\text{price}}_{\text{T}}}.
\end{align}
The final output is a compressed vector combined by $\mathbf{h}^{\text{DN}}_t$ and $\mathbf{h}^{\text{price}}_t$, summarizing both network state and market signals:
\begin{align}\label{eq:final output}
    \mathbf{h}_t^{\text{STG}} = [\mathbf{h}^{\text{DN}}_t,\mathbf{h}^{\text{price}}_t] \in \mathbb{R}^{d^{\text{DN}}_{\text{ST}} + d^{\text{price}}_{\text{T}}}.
\end{align}
The resulting joint embedding provides the DRL agent with physical dynamics and external economic incentives.
The entire data flow is parameterized as:
\begin{subequations}\label{eq:PISTGcn struc}
\begin{align}
&\omega_{\text{PISTG}}(\mathbf{X}^{\text{bus}}_{t-W+1:t};\mathbf{X}^{\text{price}}_{t-W+1:t}) = \mathbf{h}^{\text{STG}}_t,\\
&\omega_{\text{PISTG}} = \Big[\omega_{\text{GCN}}||\omega_{\text{att}}||[\omega^{\text{DN}}_{\text{TCN}} \oplus \omega^{\text{price}}_{\text{TCN}}]\Big],
\end{align}
\end{subequations}
where $\omega_{\text{PISTG}}$ is the entire network, $||$ indicates the stacking between two modules, and $\oplus$ indicates parallel concatenation.

\subsubsection{Physics-Regularized Latent Representation}\label{sec: piae}

To enhance the physical consistency of PISTG embeddings, we design a physics-informed auto-encoder (PIAE) by appending a lightweight reconstruction head $\omega_{\text{recon}}$ to the encoder $\omega_{\text{PISTG}}$. The PIAE reconstructs key physical variables, i.e., branch active/reactive flows and bus voltages $(\hat{p}_{ij,t}, \hat{q}_{ij,t}, \hat{V}_{i,t})$:
\begin{subequations}
\begin{align}
    &\omega_{\text{PIAE}}(\mathbf{X}^{\text{bus}}_t;\mathbf{X}^{\text{price}}_t;\tilde{\mathbf{A}}) = (\hat{p}_{ij,t}, \hat{q}_{ij,t}, \hat{V}_{i,t}),\\
    & \omega_{\text{PIAE}}=[\omega_{\text{PISTG}}||\omega_{\text{recon}}],
\end{align}
\end{subequations}
so that the learned embeddings align with power-system laws, improving convergence consistency and downstream decision reliability.

Since the Kirchhoff's Current Law and Ohm's law capture the most fundamental electrical constraints, we incorporate them as physics-consistency loss terms during training. The bus-level active power balance loss guides the PISTG encoder to follow KCL by aligning reconstructed flows with nodal injections and consumptions:
\begin{align}\label{eq:kcl}
\mathcal{L}_{\text{KCL}} = &\sum_{i\in\mathcal{N}} \Big \| \sum_{j:(j,i)} \hat{p}_{ij,t} + p^{\text{PV}}_{i,t} + \sum_{m:\xi(m)=i} p^{\text{dis}}_{m,t} \notag\\
& - \Big(p^{\text{load}}_{i,t} + \sum_{j:(i,j)} \hat{p}_{ij,t} + \sum_{m:\xi(m)=i} p^{\text{ch}}_{m,t}\Big) \Big\|_2^2.
\end{align}
The line-level voltage consistency term guides PISTG encoder to follow Ohm's law by constraining voltage differences across connected buses:
\begin{subequations}\label{eq:vdrop}
\begin{align}
\mathcal{L}_{\text{VDrop}} = & \sum_{(i,j)\in\mathcal{L}} \Big\|\hat{V}^2_{j,t}-\hat{V}^2_{i,t} +2\left(r_{ij}\hat{p}_{ij,t}+x_{ij}\hat{q}_{ij,t}\right) \notag\\
&-\left(r_{ij}^2+x_{ij}^2\right)\hat{\ell}_{ij,t} \Big\|_2^2,\\
&\hat{\ell}_{ij,t} = \frac{\hat{p}_{ij,t}^2+\hat{q}_{ij,t}^2}{\hat{V}^2_{i,t}}.
\end{align}
\end{subequations}

Moreover, for regularization, a reconstruction loss term validates the accuracy of the PISTG embedding. Specifically, we guide the reconstructed voltages and branch flows to match their true values by applying an MSE penalty on the predicted bus voltage, branch active power, and branch reactive power:
\begin{align}\label{eq:reconstruction}
&\mathcal{L}_{\text{recon}} = \sum_{i\in\mathcal{N}} \big|\hat{V}_{i,t} - V_{i,t}\big|_2^2+ \notag\\
&\sum_{(i,j)\in\mathcal{L}} \Big(\big|\hat{p}_{ij,t} - p_{ij,t}\big|_2^2 + \big|\hat{q}_{ij,t} - q_{ij,t}\big|_2^2\Big).
\end{align}
This reconstruction loss promotes that latent representations produced by the PISTG encoder preserve sufficient physical information to faithfully recover voltage magnitudes and power-flow patterns, thereby improving embedding fidelity and stabilizing downstream decision-making.

The overall objective jointly minimizes these three terms:
\begin{align}\label{eq:update PISTG}
 \min_{\omega_{\text{PIAE}}} \big[ \beta_1 \mathcal{L}_{\text{KCL}}(\omega_{\text{PIAE}}) + \beta_2 \mathcal{L}_{\text{VDrop}}(\omega_{\text{PIAE}}) + \beta_3 \mathcal{L}_{\text{recon}}(\omega_{\text{PIAE}})\big], 
\end{align}
where $\beta_1, \beta_2,$ and $\beta_3$ are weights. Through this, the PISTG encoder learns both compact and physically consistent embeddings.

Overall, the PI adjacency $\tilde{A}$ and the losses in Eq.~\eqref{eq:update PISTG} provide structural priors and soft training regularization for representation learning, rather than a hard power-flow feasibility layer during policy execution.

\subsection{CMDP Modeling for Problem}\label{sec:cmdp}
The coordination of CBESS units in a DN with high DER penetration can be formulated as a CMDP, where the DSO seeks to minimize operating cost while explicitly accounting for network constraints. Formally, the CMDP is defined as:
\begin{align}\label{eq: CMDP elements}
    \mathbb{M} = \langle \mathcal{S}, \mathcal{A}, \mathbb{P}, \mathcal{R}, \mathcal{C}, \gamma \rangle,
\end{align}
where $\mathcal{S}$ denotes system states embedding DN and CBESS information, $\mathcal{A}$ denotes the joint actions of CBESSs and inverters, $\mathbb{P}$ denotes the transition governed by power-flow dynamics, $\mathcal{R}$ denotes the economic-oriented reward, $\mathcal{C}$ denotes the operational-constraint costs, and $\gamma\in(0,1)$ denotes discount factor.

\subsubsection{CMDP Elements}
We formulate the CBESS coordination problem as a CMDP. At each time step, the agent observes a compact spatiotemporal state, executes coordinated active/reactive power decisions, and receives an economic reward together with multiple constraint-specific cost signals.
The key components are defined as:
\begin{subequations}
\begin{align}
    &\mathbf{s}_t = \mathbf{h}^{\text{STG}}_t \in \mathcal{S},\label{eq: s}\\
    &\mathbf{a}_t = [(p^{\text{CB}}_{m,t},q^{\text{CB}}_{m,t})_{m\in\mathcal{M}}, (\kappa_{i,t},q^{\text{inv}}_{i,t})_{i\in\mathcal{N}^{\text{DER}}}]\in \mathcal{A},\label{eq: a}\\
    &r_t = - \varphi^{\text{wh}}_t \Big[ 
      \sum_{i\in\mathcal{N}} (p^{\text{load}}_{i,t} - p^{\text{PV}}_{i,t}) - \sum_{m\in \mathcal{M}} p^{\text{CB}}_{m,t}  \Big] \Delta t , \notag \\
      &\quad - \varphi^{\text{deg}} \sum_{m\in\mathcal{M}} \big( |p^{\text{CB}}_{m,t}| \big)\Delta t, \label{eq:reward function}\\
    & c^{\text{VD}}_t = \sum_{i\in\mathcal{N}} 
    \Big[ (V_{i,t}-\overline{V})^+ + (\underline{V}-V_{i,t})^+ \Big],
    \label{eq: cost-voltage-deg} \\
    & c^{\text{VN}}_t = \sum_{i\in\mathcal{N}} \mathbb{I}\Big[ (V_{i,t} > \overline{V}) \; \lor \; (V_{i,t} < \underline{V}) \Big], \label{eq: cost-voltage-num} \\
    & c^{\text{LO}}_t = \sum_{(i,j)\in\mathcal L} \Big[\sqrt{p_{ij,t}^2 + q_{ij,t}^2} - \overline S_{ij}\Big]^+ .\label{eq: cost-line}
\end{align}
\end{subequations}
In Eq.~\eqref{eq: s}, at time $t$, the compact spatiotemporal state $\mathbf{s}_t=\mathbf{h}^{\text{STG}}_t$ encodes network- and device-level variables such as demands, PV generation, voltages, SOC, and market prices.
In Eq.~\eqref{eq: a}, the action vector $\mathbf{a}_t \in \mathcal{A}$ comprises the active $p^{\text{CB}}_{m,t}$ and reactive power $q^{\text{CB}}_{m,t}$ of each CBESS, as well as the PV curtailment ratio $\kappa_{i,t}$ and reactive power $q^{\text{inv}}_{i,t}$ of bus-level inverters on DER-rich buses.
For each CBESS, the actor outputs normalized variables $\alpha^P_{m,t},\alpha^Q_{m,t}\in[-1,1]$, which are mapped to feasible active and reactive power commands before power-flow evaluation:
\begin{align}
p^{\text{CB}}_{m,t} &= \frac{\overline{P}^{\text{dis}}_{m,t}-\overline{P}^{\text{ch}}_{m,t}}{2} + \frac{\overline{P}^{\text{dis}}_{m,t}+\overline{P}^{\text{ch}}_{m,t}}{2}\alpha^P_{m,t}, \\
q^{\text{CB}}_{m,t} &= \alpha^Q_{m,t} \sqrt{\max\left\{(S^{\text{CB}}_m)^2-(p^{\text{CB}}_{m,t})^2,0\right\}} .
\end{align}
Here, $\overline{P}^{\text{ch}}_{m,t}$ and $\overline{P}^{\text{dis}}_{m,t}$ denote the SOC-aware charging and discharging limits, respectively, and $p^{\text{CB}}_{m,t}>0$ denotes discharging. This mapping enforces charging--discharging exclusivity and satisfies Eq.~\eqref{eq: cbess thermal limits} by construction~\cite{9754695}.
In Eq.~\eqref{eq:reward function}, the reward $r_t$ represents the DSO's economic objective of minimizing operating cost. It consists of two components: (i) the energy cost, computed from the net active power and the time-varying wholesale electricity price $\varphi^{\text{wh}}_t$, and (ii) a battery degradation cost that penalizes charging and discharging throughput using a per-unit degradation coefficient $\varphi^{\text{deg}}$.
In terms of cost design, we introduce separate operational-constraint cost functions.
As for the voltage-related cost terms, we consider both voltage-violation degree (VD) and voltage-violation count (VN) to stabilize learning under rare-but-large events, since VD captures severity, while VN captures spatial frequency.
In Eq.~\eqref{eq: cost-voltage-deg}, the term $c^{\text{VD}}_t$ quantifies the aggregated magnitude of voltage deviations beyond admissible bounds across all buses.
In Eq.~\eqref{eq: cost-voltage-num}, the term $c^{\text{VN}}_t$ counts the number of buses whose voltage violates prescribed limits at time $t$.
In Eq.~\eqref{eq: cost-line}, the term $c^{\text{LO}}_t$ measures the extent of apparent power exceedance over line load limits for all distribution lines.

\subsubsection{CMDP Formulation}
Given a policy $\pi_\theta(\mathbf{a}|\mathbf{s})$, and denote the trajectory $\tau_t=\Big(\mathbf{s}_t,\mathbf{a}_t, r_t, [c^k_t]_{k \in \mathcal{K}} \Big)$, where $\mathcal{K}=\{\text{VD},\text{VN},\text{LO}\}$ contains the terms of the constraints.
We write the discounted return of the reward and costs as:
\begin{align}
&J^R(\pi_\theta) \triangleq \mathbb{E}_{\tau\sim\pi_{\theta}}\Big[\sum_{t=0}^{T-1}\gamma^t\, r_t\Big],\\
&J^C_k(\pi_\theta) \triangleq \mathbb{E}_{\tau\sim{\pi_\theta}}\Big[\sum_{t=0}^{T-1}\gamma^t \, c^k_t\Big],\quad k\in \mathcal{K}. \label{eq:j^c}
\end{align}

Then the formulation can be written as:
\begin{align}\label{eq:cmdp-discount}
\max_{\pi_\theta} \, J^R(\pi_\theta), \quad \text{s.t.} \, J^C_k(\pi_\theta) \le d_k, \, k \in \mathcal{K},
\end{align}
where $d_k$ is the expected discounted-cost budget for constraint $k$, used to regulate its Lagrangian multiplier and penalize exceedance of the corresponding budget. Throughout this paper, safe RL refers to CMDP-based constraint-aware policy learning under expected discounted cost budgets; it does not imply pointwise feasibility guarantees for every action or operating realization.

\subsection{Constraint-Specific Value Decomposition with LDCPPO}

We develop LDCPPO for CBESS coordination to optimize economic return subject to expected constraint-cost budgets. LDCPPO implements constraint-specific value decomposition through independent reward and cost critics, preserving distinct learning signals for heterogeneous voltage and line-loading risks. Adaptive Lagrangian multipliers combine these signals during actor optimization, while a shared PISTG encoder provides a common physically informed state representation for the lightweight actor and critic heads.

\subsubsection{Independent Reward and Cost Critics}

Different from a mixed-critic Lagrangian design, LDCPPO estimates the reward return and each physical cost return with separate critics, and combines their advantages only during the actor update~\cite{yang2026why}. This keeps the critic targets independent of the adaptive Lagrangian multipliers and preserves constraint-specific learning signals for voltage and line-loading control.
We employ four separate critic networks: one for the reward and three for the costs, parameterized by $\phi^R$ and $\{\phi^k\}_{k\in\mathcal{K}}$, where $k$ corresponds to the two voltage violation penalties and line-overload cost, respectively. Each critic estimates a state-value function from $\mathbf{s}_t$:
$V_{\phi^R}(\mathbf{s}_t)$ for the reward and $V_{\phi^k}(\mathbf{s}_t)$ for cost $k\in\mathcal{K}$.

\paragraph{GAE targets and advantages}
Generalized advantage estimation (GAE) is a popular way to improve PPO learning stability. First, we calculate temporal difference as:
\begin{subequations}\label{eq:TD error}
\begin{align}
\delta_t^R &= r_t + \gamma\,V_{\phi^R}(\mathbf{s}_{t+1}) - V_{\phi^R}(\mathbf{s}_t),\\
\delta_t^k &= c^k_t + \gamma\,V_{\phi^k}(\mathbf{s}_{t+1}) - V_{\phi^k}(\mathbf{s}_t),\, k\in\mathcal{K}.
\end{align}    
\end{subequations}
Next, we compute generalized advantages $A^R_t$ and $A^k_t$ (with $\lambda\!\in\![0,1]$) over truncated rollouts:
\begin{subequations}\label{eq:advantage}
\begin{align}
A^R_t &= \zeta\!\left[\sum_{l=0}^{L_t-1}(\gamma\lambda)^l\,\delta_{t+l}^R\right],\\
A^k_t &= \zeta\!\left[\sum_{l=0}^{L_t-1}(\gamma\lambda)^l\,\delta^k_{t+l}\right],\quad k\in\mathcal{K},
\end{align}
\end{subequations}
where $l$ is the dummy summation index (offset from $t$), $L_t$ is the look-ahead length from $t$, and $\zeta(\cdot)$ standardizes the advantages to zero mean and unit variance within the batch.
Accordingly, the corresponding value-regression targets:
\begin{subequations}\label{eq:value target}
\begin{align}
&\hat{R}_t = A^R_t + V_{\phi^R}(\mathbf{s}_t),\\
&\hat{C}_{k,t} = A^k_t + V_{\phi^k}(\mathbf{s}_t), \quad k\in\mathcal{K}.
\end{align}
\end{subequations}

\paragraph{Critic losses}
Given trajectories sampled from the buffer $\mathcal{B}$, we aim to minimize:
\begin{subequations}\label{Eq:critic_loss_set}
\begin{align}
\mathcal{L}^R(\phi^R) &= \mathbb{E}_{\tau\sim\mathcal{B}}\!\left[\big(V_{\phi^R}(\mathbf{s}_t)-\hat{R}_t\big)^2\right],\\
\mathcal{L}^k(\phi^k) &= \mathbb{E}_{\tau\sim\mathcal{B}}\!\left[\big(V_{\phi^k}(\mathbf{s}_t)-\hat{C}_{k,t}\big)^2\right],\quad k\in\mathcal{K}.
\end{align}
\end{subequations}

\subsubsection{Actor Network}
Actions are sampled from the policy $\pi_{\theta}$:
\begin{align}\label{Eq:action_sample}
\mathbf{a}_t \sim \pi_{\theta}(\mathbf{a}_t\mid \mathbf{s}_t).
\end{align}
To update the actor network, we need to maximize the PPO clipped surrogate:
\begin{subequations}\label{eq:actor update}
\begin{align}
&J^{\text{PPO}}(\theta) = 
\mathbb{E}_{\tau\sim\mathcal{B}} \left[
\min\Big(\rho_t(\theta) A^L_t, \right. \notag\\ 
&\left. \qquad \qquad \text{clip}(\rho_t(\theta),1-\epsilon,1+\epsilon)\,A^L_t\Big)
\right].\\
&\rho_t(\theta) = \frac{\pi_{\theta}(\mathbf{a}_t\mid \mathbf{s}_t)}{\pi_{\theta_{\text{old}}}(\mathbf{a}_t\mid \mathbf{s}_t)},\\
&A^L_t \triangleq A^R_t - \sum_{k\in\mathcal{K}} \lambda_k\,A^k_t,
\end{align}
\end{subequations}
where $\rho_t(\theta)$ is the importance ratio, $A^L_t$ is the Lagrangian advantage computed by summing the reward GAE and cost GAEs using Lagrange multipliers $\{\lambda_k\}_{k\in\mathcal{K}}$.

\subsubsection{Parameter Update}

For each episode or mini-batch, perform gradient descent for critics/actor and projected dual ascent for multipliers.
Additionally, the Lagrangian multipliers $\{\lambda^k\}_{k\in\mathcal{K}}$ are adaptively updated through dual ascent to regulate the expected constraint-cost budgets, maintaining their non-negativity and adjusting based on the deviation of constraint values from their respective thresholds $d^k$:
\begin{subequations}\label{Eq:agent_update}
\begin{align}
\phi^R &\leftarrow \phi^R - \eta^R      \nabla_{\phi^R}\, \mathcal{L}^R,\\
\phi^k    &\leftarrow \phi^k    - \eta^k    \nabla_{\phi^k}\, \mathcal{L}^k,\quad k\in\mathcal{K},\\
\theta      &\leftarrow \theta      - \eta^{\text{Act}} \nabla_{\theta}\, \mathcal{L}_{\theta},\\
\lambda^k   &\leftarrow [\lambda^k+\eta^{\text{Lag}}(\hat{J}^C_k-d^k)]^{+}, \forall k \in \mathcal{K},
\end{align}
\end{subequations}
where $\nabla$ is the gradient descent operation, and $\eta^{\text{R}}$, $\{\eta^{k}\}_{k\in\mathcal{K}}$, $\eta^{\text{Act}}$, $\eta^{\text{Lag}}$ are the learning rates for different updates. Note that $\hat{J}^C_k$ is the minibatch mean of discounted cost returns $J^C_k$ in Eq.~\eqref{eq:j^c}, and $[\cdot]^{+}$ enforces $\lambda^k \geq 0$.

\begin{figure}[t]
\centering
\includegraphics[width=0.45\textwidth]{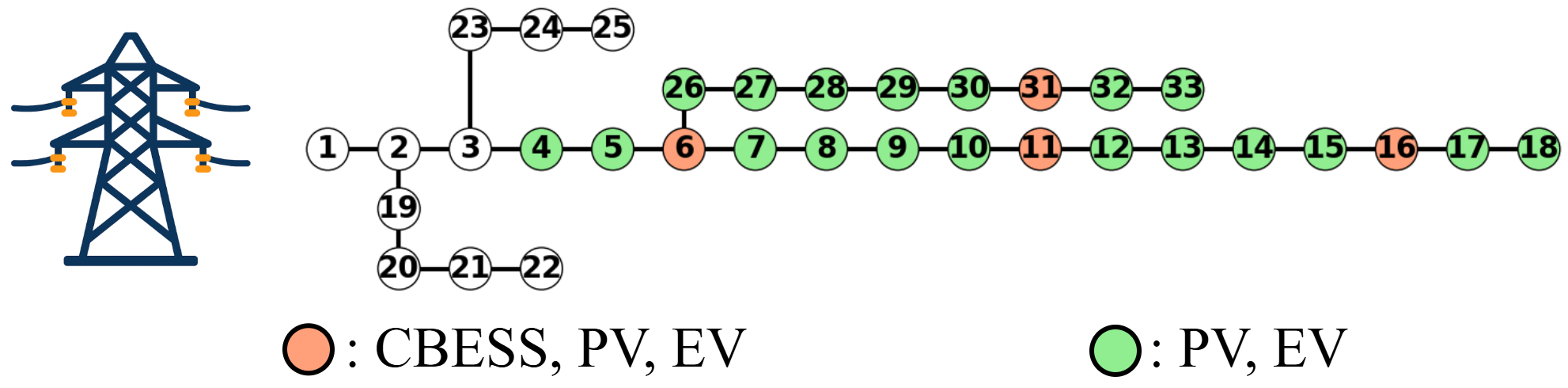}
\caption{CBESS deployment in the IEEE 33-bus system with high DER penetration.} 
\label{fig:33bus}
\end{figure}

\begin{figure}[!t]
\centering
\renewcommand{\subfigcapskip}{-5pt}
    \subfigure[Daily PV and load profile]{
    \includegraphics[width=.23\textwidth]{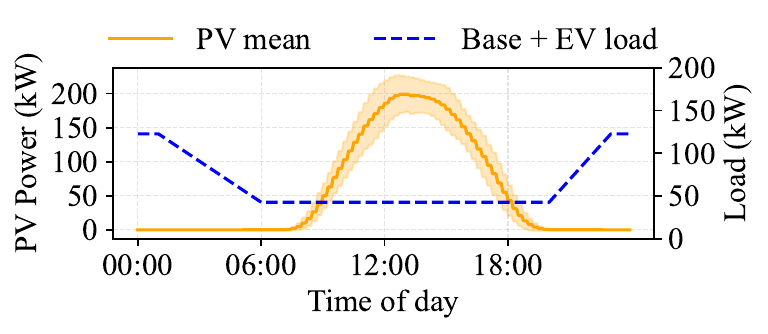}  }
    \hspace{-0.05\linewidth}
\renewcommand{\subfigcapskip}{-5pt}
    \subfigure[Aggregated energy scale comparison between one bus and one CBESS]{
    \includegraphics[width=.23\textwidth]{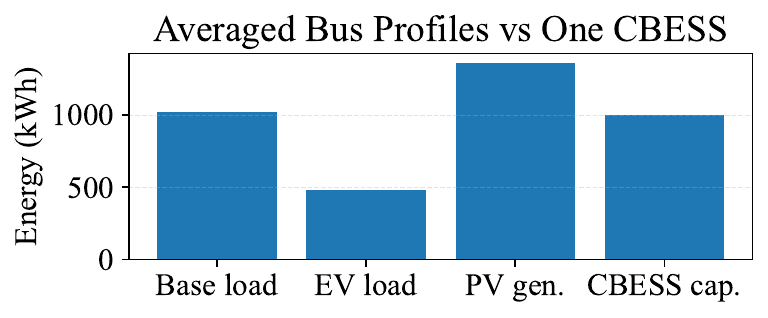}  }

\caption{System setup of PV, EV, load, and CBESSs.}
\label{fig:profiles}
\end{figure}

\section{Simulations}\label{Sec:E}

\subsection{Experimental Setup}

\begin{table}[!t]
\centering
\caption{Hyperparameter and network architecture settings.}
\label{tab:exp_hyperparameters}
\scriptsize
\renewcommand\arraystretch{1.0}
\setlength{\tabcolsep}{1.5mm}
\begin{tabularx}{\columnwidth}{p{2cm} p{4cm} p{1.15cm}}
\toprule
Category & Parameter & Value \\
\midrule

\multirow{10}{*}{PISTG encoder}
& PI adjacency constant, $\epsilon$ & $10^{-6}$ \\
& Capacity-weight coefficient, $\nu$ & 1.0 \\
& GCN layers, $L$ & 2 \\
& GCN hidden dimension & 64 \\
& Spatial embedding dim. $d^S$ & 64 \\
& DN-TCN layers & 2 \\
& Price-TCN layers & 2 \\
& TCN kernel size & 3 \\
& TCN dilations & $\{1,2\}$ \\
& Joint embedding dim. $d^{\mathrm{STG}}$ & 80 \\
\midrule

\multirow{5}{*}{PISTG pre-training}
& PIAE decoder hidden layers & $[128,128]$ \\
& KCL loss weight, $\beta_1$ & 1.0 \\
& Voltage-drop loss weight, $\beta_2$ & 1.0 \\
& Reconstruction loss weight, $\beta_3$ & 0.5 \\
& PIAE learning rate & $1\times 10^{-3}$ \\
\midrule

\multirow{10}{*}{LDCPPO}
& Actor hidden layers & $[128,128]$ \\
& Critic hidden layers & $[128,128]$ \\
& Discount factor, $\gamma$ & 0.99 \\
& GAE parameter, $\lambda_{\mathrm{GAE}}$ & 0.95 \\
& PPO clip range, $\epsilon_{\mathrm{clip}}$ & 0.2 \\
& Target KL & 0.015 \\
& Actor learning rate, $\eta^{\mathrm{Act}}$ & $3\times10^{-5}$ \\
& Critic learning rate, $\eta^R,\eta^k$ & $3\times10^{-5}$ \\
& Dual learning rate, $\eta^{\mathrm{Lag}}$ & $5\times10^{-4}$ \\
& Initial multipliers, $\lambda^k_0$ & 0 \\

\bottomrule
\end{tabularx}
\end{table}

\subsubsection{Simulator Setup}

The case study is based on the IEEE 33-bus system shown in Fig. \ref{fig:33bus}.
The slack bus fixed at 1 p.u., and bus voltage threshold fixed within [0.95,1.05] p.u.
Four CBESS units are placed at buses 6, 11, 16, and 31, each of a capacity of 1000 kWh, an initial SOC of 50\%, a charge/discharge rate limit of [-250,250] kW, a reactive power range of [-50,50] kVAr, charge/discharge efficiency of 0.95, and battery degradation cost of 0.03 AUD per kWh~\cite{battery_cost}.
The operation horizon is one day with 288 steps (5 min each). 
Accordingly, the actions in Eq.~\eqref{eq: a} are updated every 5 min and held until the next step, while faster intra-interval variations are assumed to be handled by local inverter-based Volt--VAR control and existing protection functions, and are not explicitly captured by the 5-min constraint evaluation.

We set DER buses around these CBESSs. 
We construct the 5-min PV generation and load profiles on each bus by linear interpolation of the Solar Home Electricity Data provided by Ausgrid~\cite{ratnam2017residential}, which comprises three years of half-hourly electricity data for 300 randomly selected solar homes.
Moreover, to simulate household consumption peak, we add residential EV charging power for each DER bus over night (8:00 PM to 6:00 AM).
The profiles are demonstrated in Fig.~\ref{fig:profiles}.

The main hyperparameter and network architecture settings of the PISTG encoder, PISTG pre-training, and LDCPPO are summarized in Table~\ref{tab:exp_hyperparameters}.
The data scale is also specified to improve reproducibility. With a 5-min control interval, each daily episode contains $T=288$ decision steps, and 2000 training episodes provide approximately $5.76\times10^5$ state--action--reward--cost transition samples. The rolling window length $W=12$ used by the PISTG encoder further gives about $5.54\times10^5$ graph-temporal training windows.

\subsubsection{Comparison Baselines \& Computational details}

We compare our method with seven DRL baselines: PPO, TD3, SAC, LDCPPO, STG-LDCPPO, PISTG-CPO, and PI-GR-STG-LDCPPO.
PPO, TD3, and SAC are classical DRL algorithms widely adopted in continuous control tasks.
LDCPPO and STG-LDCPPO are included as closely related constrained baselines to isolate the effects of Lagrangian-based constraint handling, graph-based state encoding, and physics-informed training within the proposed framework.
PISTG-CPO combines the proposed PISTG encoder with a classical constrained policy optimization update~\cite{achiam2017constrained,10923740}, while PI-GR-STG-LDCPPO adapts the physics-informed graphical representation design~\cite{10113230} to the same STG-LDCPPO framework.
For PPO, TD3, and SAC, constraint costs are aggregated into the reward via weighted penalties, resulting in a single scalar objective.
In contrast, LDCPPO, STG-LDCPPO, PI-GR-STG-LDCPPO, and the proposed method adopt a Lagrangian relaxation framework with constraint-specific critics, while PISTG-CPO represents a classical constrained policy optimization baseline.

Notably, optimization-based approaches were not included as baselines in this work, which was a choice of scope, which focuses on improving utilities while satisfying multiple constraints by learning-based policies, rather than a missing consideration.
Also, related work~\cite{9680706,10288542} has shown learning-based methods can remain competitive while offering much faster online control, showing the promise of RL methods. 
Our work aims to evaluate the capability of dealing with key challenges in learning-based approaches by contributing to novel physics-informed representation and constraint-specific critics, which have been demonstrated to be effective.

Training uses a learning rate of 3e-5, clipping of 0.2, target KL of 0.015, value coefficient of 0.5, zero entropy bonus, gradient norm clipping of 0.5, and Gaussian log-std initialized at -0.5, with $\gamma$ set to 0.99, $\lambda^k$ set to 0.95, dual learning rate of 5e-4, over 2000 episodes of 288 steps (5-min resolution).
All data collection processes and training are conducted on a server with 8 NVIDIA GeForce 3090 GPUs, each of which has 24G memory. The CUDA version is 11.5.

\subsection{Result Analysis}

\begin{figure}[!t]
\centering
\includegraphics[width=0.9\linewidth]{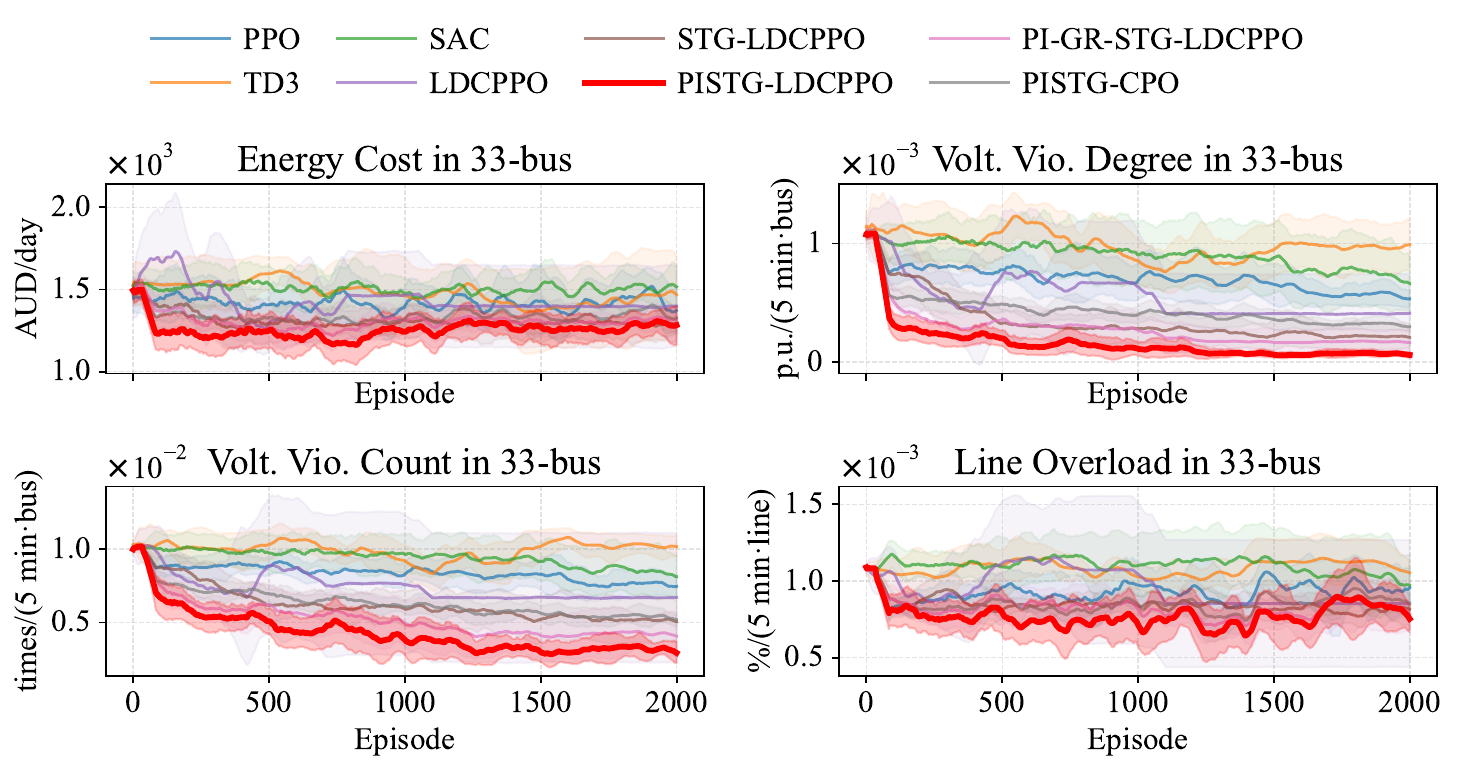}
\caption{Training performance on the four key metrics of the compared methods on the IEEE 33-bus system. The curves show the mean performance over episodes with shaded regions indicating variance across multiple runs.}
\label{fig:training curves}
\end{figure}

\begin{table}[!t]
\centering
\scriptsize
\caption{Test performance comparison over 100 episodes for 5 runs.}
\label{tab:test_results}
\renewcommand\arraystretch{1.0}
\setlength{\tabcolsep}{1.4mm}
\begin{tabular}{|p{1cm}|p{1.3cm}|p{1.8cm}|p{1.8cm}|p{1.5cm}|}
\hline
Method
& Operating cost
& Volt. vio. degree
& Volt. vio. count
& Line-overload degree \\
& AUD/day
& p.u./$(5$min$\cdot$bus$)$
& times/$(5$min$\cdot$bus$)$
& \%/$(5$min$\cdot$line$)$ \\
\hline
TD3 
& 1583.7 $\pm$ 68.4 
& (1.63 $\pm$ 0.27) $\times 10^{-3}$
& (1.02 $\pm$ 0.16) $\times 10^{-2}$
& (1.93 $\pm$ 0.39) $\times 10^{-3}$\\
\hline

SAC 
& 1546.2 $\pm$ 69.1 
& (1.52 $\pm$ 0.23) $\times 10^{-3}$
& (0.94 $\pm$ 0.14) $\times 10^{-2}$
& (1.82 $\pm$ 0.36) $\times 10^{-3}$\\
\hline

PPO
& $1424.5 \pm 53.7$
& $(1.28 \pm 0.19)\times 10^{-3}$
& $(0.87 \pm 0.11)\times 10^{-2}$
& $(1.54 \pm 0.28)\times 10^{-3}$ \\ \hline
LDCPPO
& $1453.1 \pm 49.8$
& $(0.77 \pm 0.15)\times 10^{-3}$
& $(0.79 \pm 0.10)\times 10^{-2}$
& $(1.32 \pm 0.21)\times 10^{-3}$ \\ \hline
STG-LDCPPO
& $1414.8 \pm 36.1$
& $(0.53 \pm 0.09)\times 10^{-3}$
& $(0.63 \pm 0.06)\times 10^{-2}$
& $(1.15 \pm 0.14)\times 10^{-3}$ \\\hline
PISTG-CPO
& $1420.0 \pm 37.4$
& $(0.70 \pm 0.08)\times 10^{-3}$
& $(0.63 \pm 0.05)\times 10^{-2}$
& $(1.35 \pm 0.16)\times 10^{-3}$ \\\hline
PI-GR-STG-LDCPPO
& $1412.3 \pm 42.1$
& $(0.49 \pm 0.12)\times 10^{-3}$
& $(0.52 \pm 0.12)\times 10^{-2}$
& $(1.14 \pm 0.19)\times 10^{-3}$ \\\hline
\textbf{PISTG-LDCPPO}
& $\textbf{1382.4} \pm \textbf{38.9}$
& $\textbf{(0.25} \pm \textbf{0.04)}\times 10^{-3}$
& $\textbf{(0.37} \pm \textbf{0.05)}\times 10^{-2}$
& $\textbf{(1.13} \pm \textbf{0.14)}\times 10^{-3}$ \\
\hline
\end{tabular}
\end{table}

\subsubsection{Overall Performance}

Figure~\ref{fig:training curves} and Table~\ref{tab:test_results} present the training- and test-phase performance comparison of eight methods in the CBESS coordination problem, respectively.
Four metrics are evaluated throughout the training process: 1) operating cost, which represents the economic objective; 2) voltage-violation degree, which measures the magnitude of voltage-limit exceedance at each bus and time step; 3) voltage-violation count, which measures the frequency of bus-voltage violations; and 4) line-overload degree, which measures line-capacity exceedance at each time step.

\begin{figure}[!t]
\centering
\includegraphics[width=0.9\linewidth]{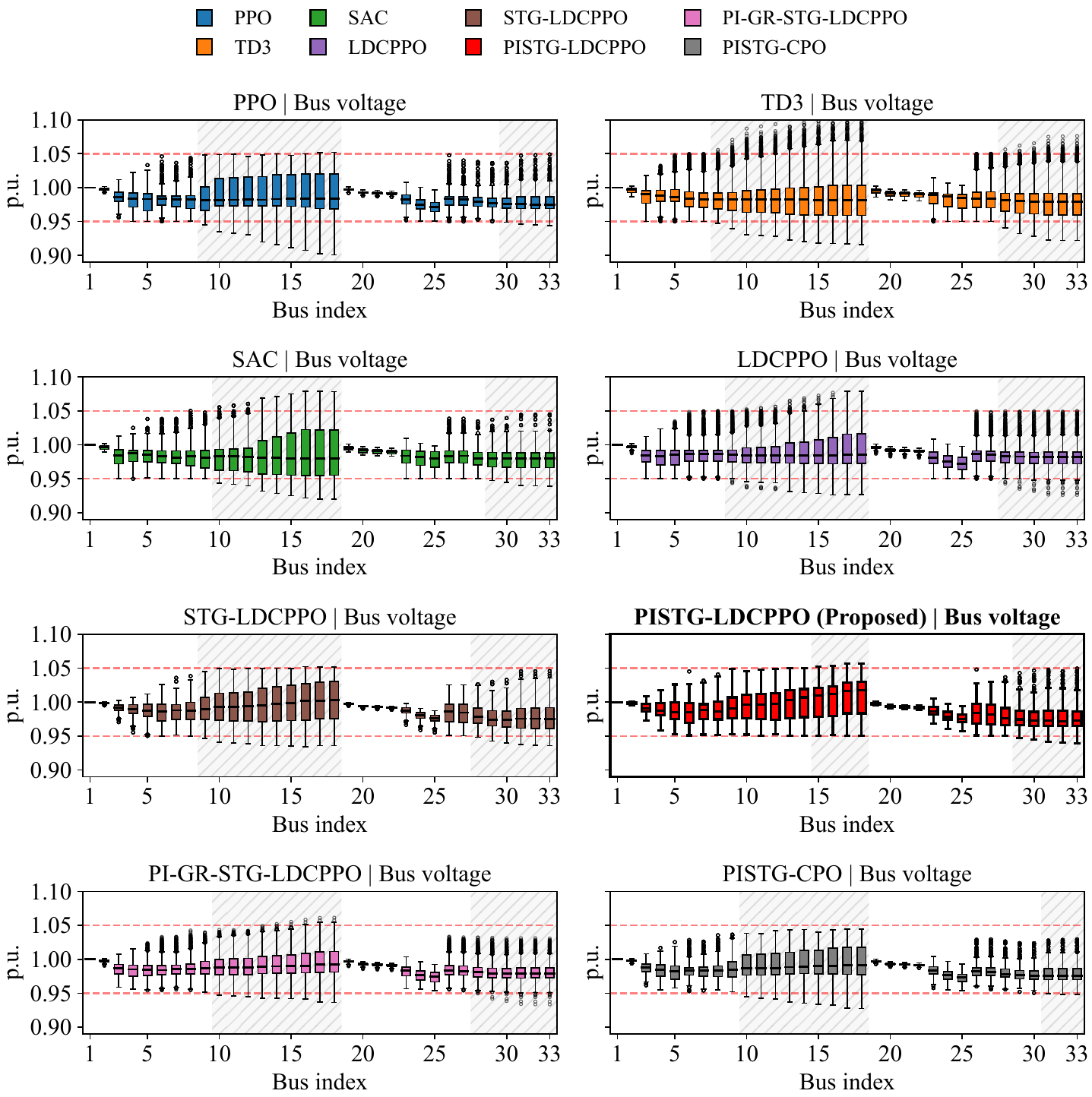}
\caption{Bus voltage distribution in test.}
\label{fig:voltage boxplot}
\end{figure}

\begin{figure}[!t]
\centering
\includegraphics[width=0.9\linewidth]{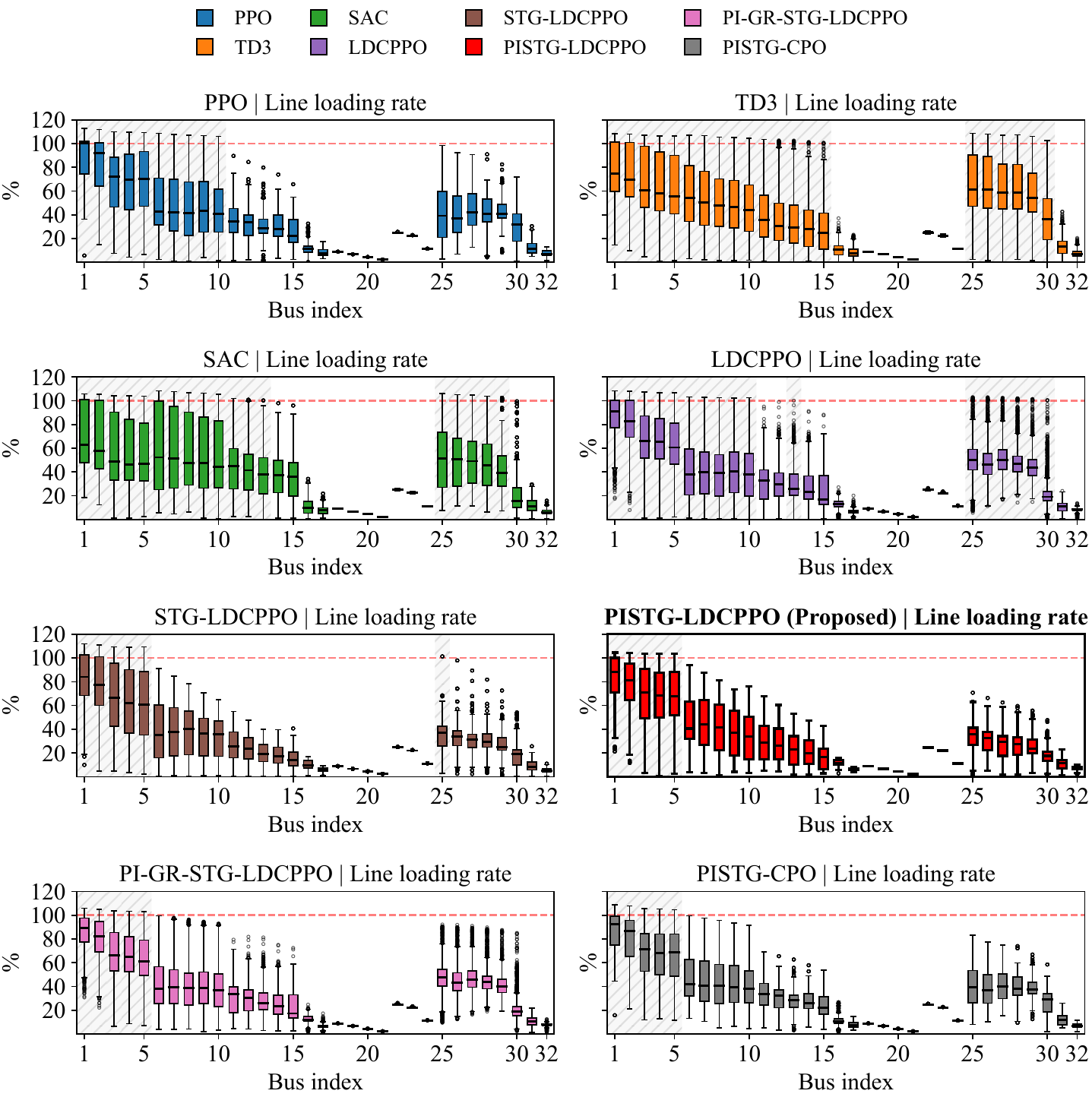}
\caption{Line loading rate distribution in test.}
\label{fig:ll boxplot}
\end{figure}

Among vanilla DRL methods, PPO, SAC and TD3 converge with non-negligible steady-state voltage violations and persistent line-overload levels. Introducing LDCPPO reduces empirical constraint violations in terms of violation magnitude, frequency, and line-overload degree. Furthermore, incorporating STG-LDCPPO further suppresses these violations. 
The PISTG-CPO and PI-GR-STG-LDCPPO further show that either using a classical constrained update with PISTG or adding prior physics-informed graph regularization to STG is helpful but insufficient, because effective CBESS coordination requires both constraint-specific value estimation and physically consistent spatiotemporal embeddings.
Overall, PISTG-LDCPPO achieves the lowest empirical constraint violations with competitive operating cost. Compared with all baselines in Table~\ref{tab:test_results}, it reduces the voltage-violation degree by approximately $49.0\%$--$80.5\%$, the voltage-violation count by $28.8\%$--$57.5\%$, the line-overload degree by $0.9\%$--$26.6\%$, and the operating cost by about $2.1\%$--$4.9\%$.
Together, these results confirm the complementary roles of PISTG and constraint-specific critics. Their combination achieves the strongest reduction in heterogeneous network-constraint violations, supporting the effectiveness of the proposed problem-driven integration.

\begin{figure}[!t]
\centering
\renewcommand{\subfigcapskip}{-5pt}
    \subfigure[min. and max. bus voltage]{
    \includegraphics[width=.23\textwidth]{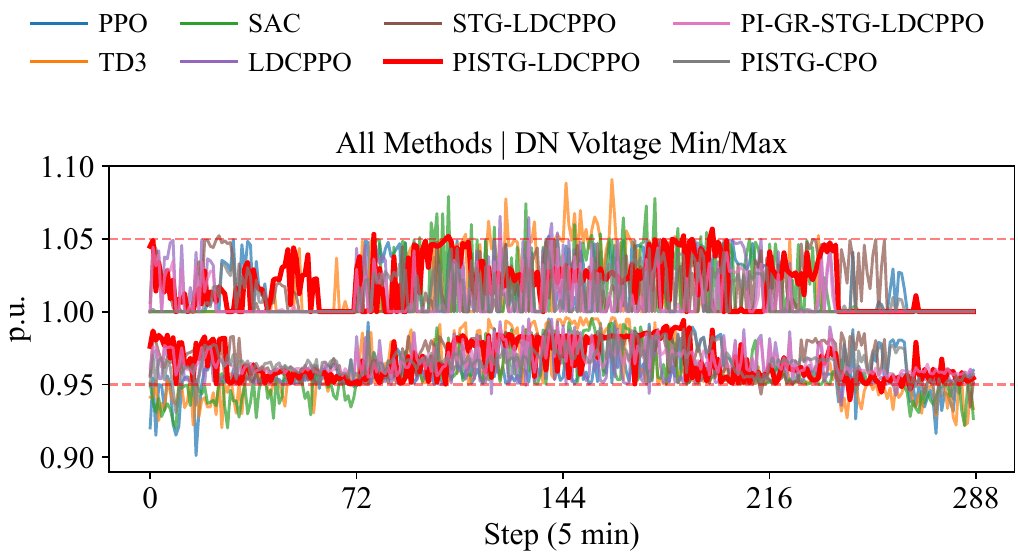}  }
    \hspace{-0.05\linewidth}
\renewcommand{\subfigcapskip}{-5pt}
    \subfigure[max. line loading rate]{
    \includegraphics[width=.23\textwidth]{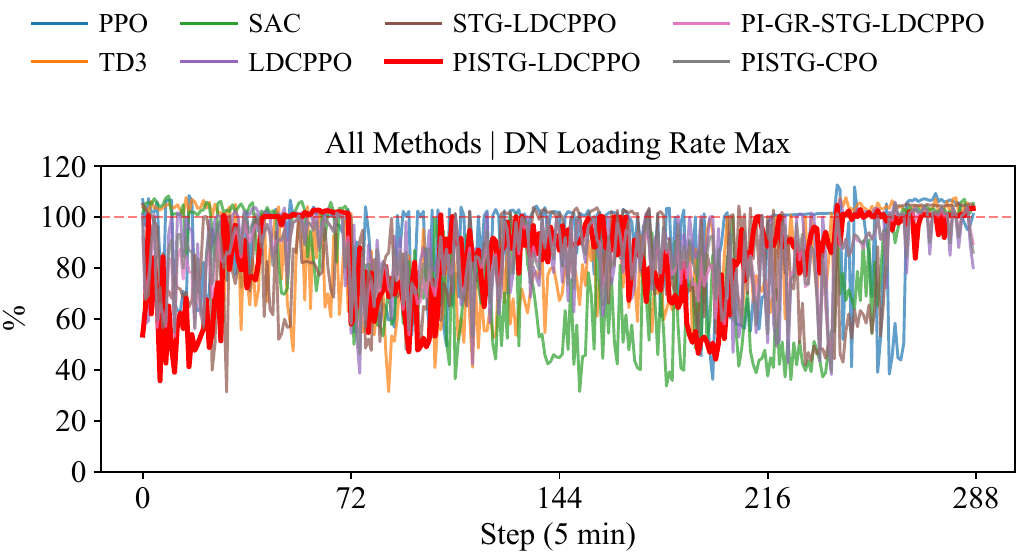}  }
\caption{Average daily evolution of system-level safety indicators.}
\label{fig:minmax curve}
\end{figure}

\begin{figure}[!t]
\centering
\renewcommand{\subfigcapskip}{-5pt}
\includegraphics[width=.49\textwidth]{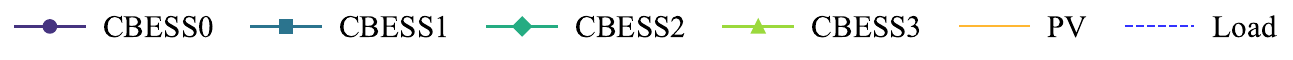}
\renewcommand{\subfigcapskip}{-5pt}
    \subfigure{
    \includegraphics[width=.23\textwidth]{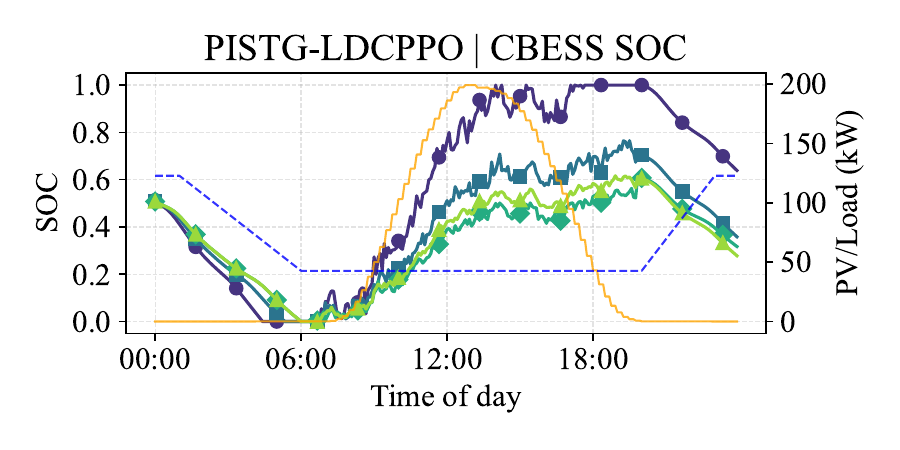}  }
    \hspace{-0.05\linewidth}
\renewcommand{\subfigcapskip}{-5pt}
    \subfigure{
    \includegraphics[width=.23\textwidth]{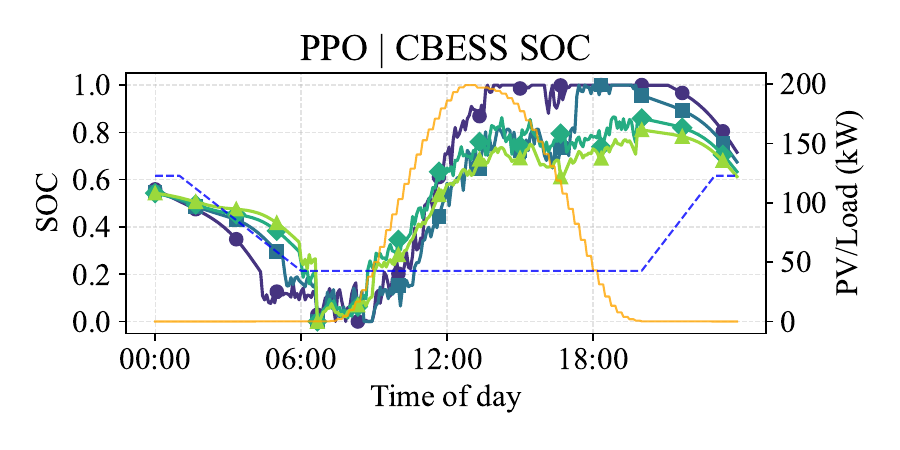}  }
\caption{Daily SOC trajectories of four CBESS units under PISTG-LDCPPO and PPO during testing.}
\label{fig:cbess curves}
\end{figure}

\subsubsection{Analysis of Voltage and Load}

The bus voltage in Fig.~\ref{fig:voltage boxplot} and line loading rate boxplots in Fig.~\ref{fig:ll boxplot} jointly illustrate the distribution of network states across buses and lines over the test horizon, with dashed lines marking the operational limits. The shaded regions further highlight the violating ranges, while the plotted outliers indicate extreme operating states beyond the main distribution.
Compared with the wider dispersion and longer tails observed in the vanilla baselines, the proposed methods progressively tighten both voltage and loading distributions. PISTG-LDCPPO produces the most compact and well-confined distributions across buses and lines, with fewer extreme values and reduced variability.

The averaged evolutions over 100 test episodes of DN-wide extremes, i.e., minimum/maximum bus voltage in Fig.~\ref{fig:minmax curve}(a) and maximum line loading in Fig.~\ref{fig:minmax curve}(b), provide a system-level view of safety during operation.
Compared to the baselines that show larger oscillations and more pronounced peaks/troughs in these extreme trajectories, PISTG-LDCPPO's voltage extremes are better confined within the admissible band and the maximum loading trajectory violates less from the line load threshold with reduced fluctuation.

Figure~\ref{fig:cbess curves} compares the daily SOC trajectories of the four CBESS units under PISTG-LDCPPO and PPO, alongside the PV and load profiles. 
During the daytime PV peak, PPO tends to charge aggressively and saturate early, whereas PISTG-LDCPPO adopts a more moderated charging strategy that better preserves scheduling flexibility. During the evening peak, PISTG-LDCPPO discharges more proactively to support network operation, with only a slight sacrifice in economic performance.
This behavior reflects a more grid-aware dispatch strategy that prioritizes network support and lower operational-constraint violations.

\subsubsection{Scalability Analysis on the IEEE 69-Bus System}
\label{subsec:scalability_69bus}

\begin{figure}[!t]
\centering
\includegraphics[width=0.86\linewidth]{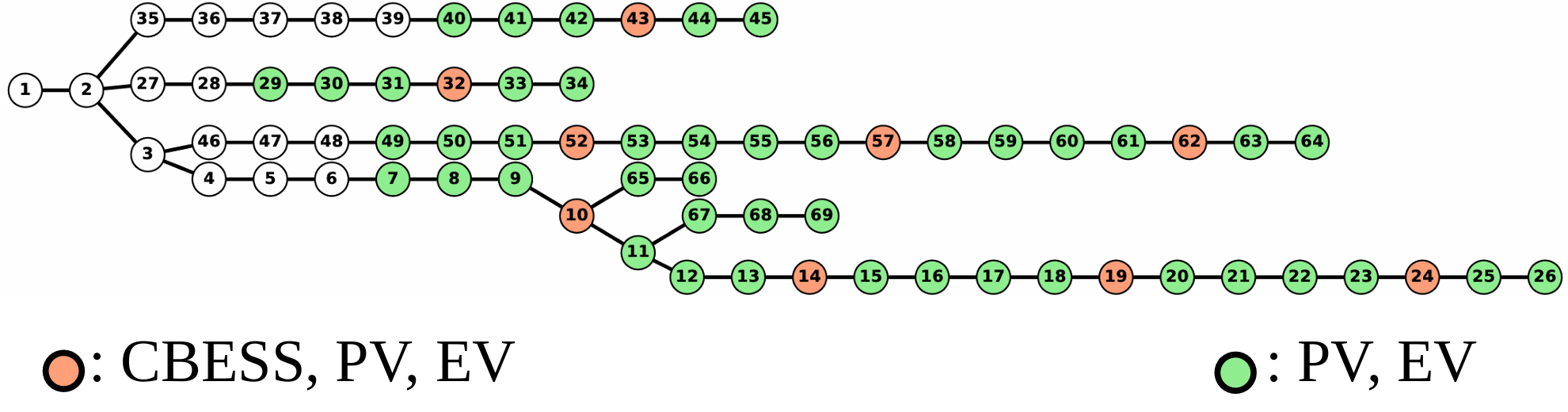}
\caption{CBESS deployment in the IEEE 69-bus system with high DER penetration.}
\label{fig:69_bus}
\end{figure}

\begin{figure}[!t]
\centering
\includegraphics[width=0.9\linewidth]{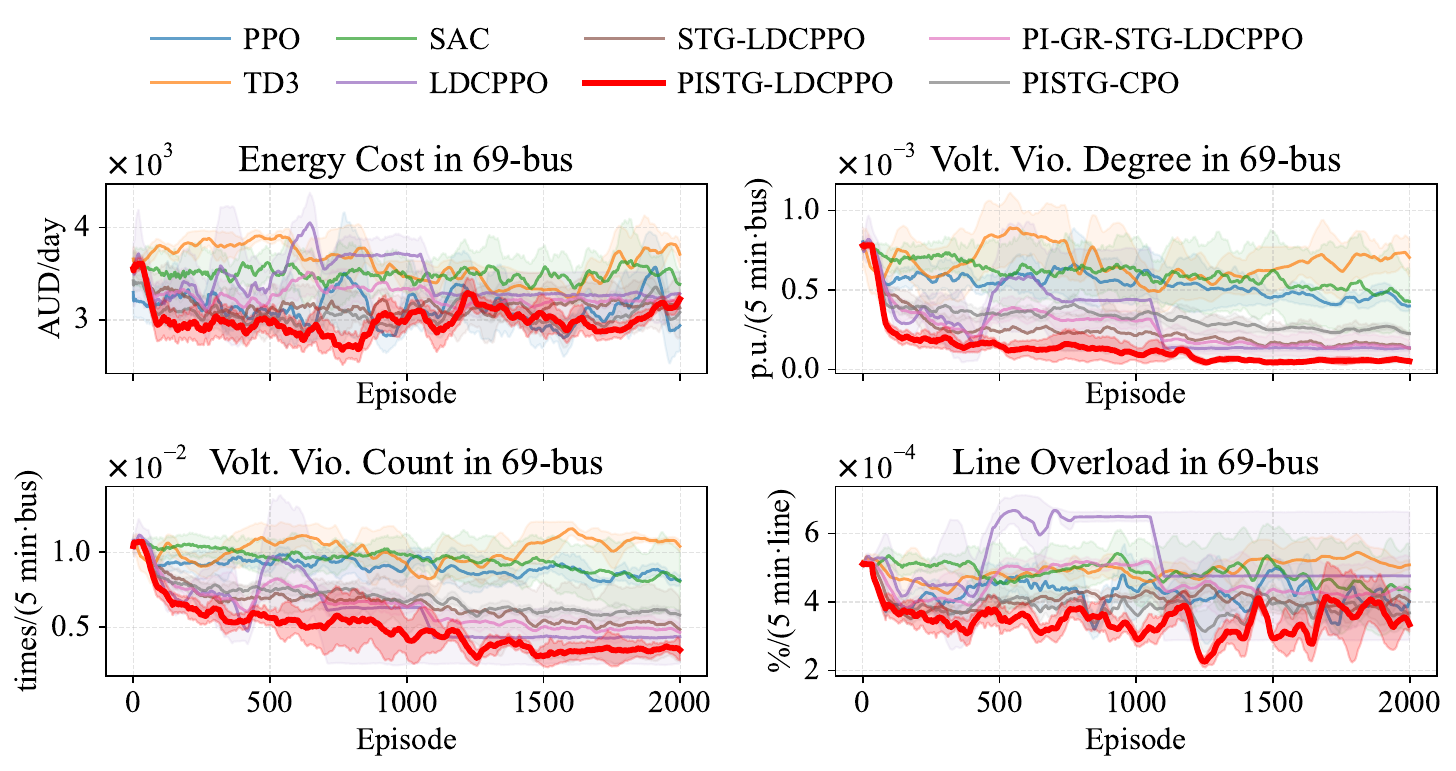}
\caption{Training performance on the four key metrics of the compared methods on IEEE 69-bus system. The curves show the mean performance over episodes with shaded regions indicating variance across multiple runs.}
\label{fig:69_training_curves}
\end{figure}

\begin{table}[!t]
\centering
\scriptsize
\caption{Test performance comparison over 100 episodes for 5 runs on IEEE 69-bus system.}
\renewcommand\arraystretch{1.0}
\setlength{\tabcolsep}{1.4mm}
\begin{tabular}{|p{1cm}|p{1.3cm}|p{1.8cm}|p{1.8cm}|p{1.5cm}|}
\hline
Method
& Operating Cost (AUD/day)
& Volt. Vio. Degree [p.u./(5min$\cdot$bus)]
& Volt. Vio. Count [times/(5min$\cdot$bus)]
& Line-Overload Degree [\%/(5min$\cdot$line)]\\
\hline
TD3
& 3740.8 $\pm$ 190.4
& (1.98 $\pm$ 0.42) $\times 10^{-3}$
& (2.28 $\pm$ 0.35) $\times 10^{-2}$
& (1.85 $\pm$ 0.42) $\times 10^{-3}$\\ \hline

SAC
& 3455.6 $\pm$ 210.7
& (2.12 $\pm$ 0.48) $\times 10^{-3}$
& (1.91 $\pm$ 0.43) $\times 10^{-2}$
& (1.92 $\pm$ 0.35) $\times 10^{-3}$\\\hline

PPO
& 2965.2 $\pm$ 245.6
& (1.60 $\pm$ 0.22) $\times 10^{-3}$
& (1.22 $\pm$ 0.20) $\times 10^{-2}$
& (1.76 $\pm$ 0.25) $\times 10^{-3}$\\\hline

LDCPPO
& 3265.4 $\pm$ 260.1
& (1.03 $\pm$ 0.18) $\times 10^{-3}$
& (0.84 $\pm$ 0.14) $\times 10^{-2}$
& (1.43 $\pm$ 0.15) $\times 10^{-3}$\\\hline

STG-LDCPPO
& 3135.5 $\pm$ 95.4
& (0.82 $\pm$ 0.15) $\times 10^{-3}$
& (0.76 $\pm$ 0.18) $\times 10^{-2}$
& (1.22 $\pm$ 0.12) $\times 10^{-3}$\\\hline
PI-GR-STG-LDCPPO
& 3122.1 $\pm$ 172.7
& (0.67 $\pm$ 0.12) $\times 10^{-3}$
& (0.67 $\pm$ 0.11) $\times 10^{-2}$
& (1.23 $\pm$ 0.14) $\times 10^{-3}$\\\hline

PISTG-CPO
& 3013.0 $\pm$ 190.4
& (1.06 $\pm$ 0.16) $\times 10^{-3}$
& (0.91 $\pm$ 0.15) $\times 10^{-2}$
& (1.46 $\pm$ 0.20) $\times 10^{-3}$\\\hline
\textbf{PISTG-LDCPPO}
& \textbf{3060.7} $\pm$ \textbf{135.2}
& \textbf{(0.52} $\pm$ \textbf{0.09)} $\times 10^{-3}$
& \textbf{(0.60} $\pm$ \textbf{0.10)} $\times 10^{-2}$
& \textbf{(1.15} $\pm$ \textbf{0.14)} $\times 10^{-3}$\\
\hline
\end{tabular}
\label{tab:69_test}
\end{table}

\begin{table}[!t]
\centering
\caption{Inference time and parameter count on IEEE 33-bus and 69-bus systems.}
\label{tab:inference_params}
\scriptsize
\setlength{\tabcolsep}{1.0mm}
\renewcommand{\arraystretch}{1.0}
\resizebox{\columnwidth}{!}{
\begin{tabular}{lcccc}
\hline
\multirow{2}{*}{Method}
& \multicolumn{2}{c}{33-bus}
& \multicolumn{2}{c}{69-bus}\\
\cline{2-5}
& Time & Params
& Time & Params\\
& (ms) & (M)
& (ms) & (M)\\
\hline
PPO
& \textcolor{gray!75}{\rule{0.36cm}{5pt}} 0.34
& \textcolor{gray!75}{\rule{0.36cm}{5pt}} 0.10
& \textcolor{gray!75}{\rule{0.45cm}{5pt}} 0.42
& \textcolor{gray!75}{\rule{0.41cm}{5pt}} 0.12\\
LDCPPO
& \textcolor{gray!75}{\rule{0.52cm}{5pt}} 0.49
& \textcolor{gray!75}{\rule{0.82cm}{5pt}} 0.24
& \textcolor{gray!75}{\rule{0.62cm}{5pt}} 0.58
& \textcolor{gray!75}{\rule{0.96cm}{5pt}} 0.28\\
STG-LDCPPO
& \textcolor{gray!75}{\rule{1.26cm}{5pt}} 1.18
& \textcolor{gray!75}{\rule{1.61cm}{5pt}} 0.47
& \textcolor{gray!75}{\rule{1.85cm}{5pt}} 1.74
& \textcolor{gray!75}{\rule{1.82cm}{5pt}} 0.53\\
\textbf{PISTG-LDCPPO}
& \textcolor{blue!70!black}{\rule{1.30cm}{5pt}} \textbf{1.22}
& \textcolor{blue!70!black}{\rule{1.61cm}{5pt}} \textbf{0.47}
& \textcolor{blue!70!black}{\rule{1.90cm}{5pt}} \textbf{1.78}
& \textcolor{blue!70!black}{\rule{1.82cm}{5pt}} \textbf{0.53}\\
\hline
\end{tabular}
}
\end{table}

To examine scalability, we further test the proposed framework on the IEEE 69-bus distribution system, as shown in Fig.~\ref{fig:69_bus}. Compared with the 33-bus case, the number of CBESS units increases from 4 to 9; PV is deployed at 53 of the 68 non-slack buses, giving a PV deployment ratio of $77.9\%$, and the PV penetration level is kept at approximately $90\%$.

Figure~\ref{fig:69_training_curves} and Table~\ref{tab:69_test} evaluate the proposed framework on the IEEE 69-bus system, where PISTG-LDCPPO maintains consistent performance and achieves lower empirical network-constraint violations than the baseline DRL and safe-DRL methods. Table~\ref{tab:inference_params} further shows that its inference time increases only from $1.22$ ms on the 33-bus system to $1.78$ ms on the 69-bus system, with the parameter count increasing from $0.47$ M to $0.53$ M, remaining negligible relative to the 5-min control interval. Overall, these results demonstrate that PISTG-LDCPPO scales effectively to larger distribution networks while maintaining constraint-aware performance and low computational overhead.

\subsubsection{PV Curtailment Analysis}

We evaluate PV curtailment test-time behavior by reporting the curtailed PV energy $E^{\mathrm{curt}}$, the PV curtailment ratio $\rho^{\mathrm{curt}}$, and the high-PV curtailment share $\eta^{\mathrm{HP}}$:
\begin{subequations}
\begin{align}
E^{\mathrm{curt}} &= \sum_{t\in\mathcal{T}}\sum_{i\in\mathcal{N}^{\mathrm{DER}}}
\kappa_{i,t}\tilde{p}^{\mathrm{PV}}_{i,t}\Delta t,\\
\rho^{\mathrm{curt}} &= 
\frac{\sum_{t\in\mathcal{T}}\sum_{i\in\mathcal{N}^{\mathrm{DER}}}
\kappa_{i,t}\tilde{p}^{\mathrm{PV}}_{i,t}\Delta t}
{\sum_{t\in\mathcal{T}}\sum_{i\in\mathcal{N}^{\mathrm{DER}}}
\tilde{p}^{\mathrm{PV}}_{i,t}\Delta t}\times 100\%,\\
\eta^{\mathrm{HP}} &=
\frac{\sum_{t\in\mathcal{T}^{\mathrm{HP}}}\sum_{i\in\mathcal{N}^{\mathrm{DER}}}
\kappa_{i,t}\tilde{p}^{\mathrm{PV}}_{i,t}\Delta t}
{\sum_{t\in\mathcal{T}}\sum_{i\in\mathcal{N}^{\mathrm{DER}}}
\kappa_{i,t}\tilde{p}^{\mathrm{PV}}_{i,t}\Delta t}\times 100\%,
\end{align}
\end{subequations}
where $\mathcal{T}^{\mathrm{HP}}$ denotes the high-PV period. The first two metrics measure the absolute and normalized curtailment levels, while $\eta^{\mathrm{HP}}$ indicates whether curtailment is mainly concentrated during high-PV periods.

\begin{table}[t]
\centering
\scriptsize
\caption{PV curtailment statistics during testing.}
\label{tab:pv_curtailment}
\renewcommand\arraystretch{1.0}
\setlength{\tabcolsep}{4.0pt}
\begin{tabular}{|l|c|c|c|c|}
\hline
\textbf{Method} & \textbf{Case} & 
\textbf{$E^{\mathrm{curt}}$} \textbf{(kWh/day)} & 
\textbf{$\rho^{\mathrm{curt}}$} \textbf{(\%)} & 
\textbf{$\eta^{\mathrm{HP}}$} \textbf{(\%)} \\
\hline
\multirow{2}{*}{TD3}
& 33 & $142.6{\pm}19.8$ & $3.57{\pm}0.50$ & $56.8$\\
& 69 & $386.4{\pm}42.7$ & $4.55{\pm}0.50$ & $55.3$\\
\hline
\multirow{2}{*}{SAC}
& 33 & $134.1{\pm}18.5$ & $3.35{\pm}0.46$ & $59.4$\\
& 69 & $338.7{\pm}39.6$ & $3.98{\pm}0.47$ & $59.1$\\
\hline
\multirow{2}{*}{PPO}
& 33 & $111.3{\pm}15.6$ & $2.78{\pm}0.39$ & $63.5$\\
& 69 & $226.9{\pm}28.4$ & $2.67{\pm}0.33$ & $65.8$\\
\hline
\multirow{2}{*}{LDCPPO}
& 33 & $104.6{\pm}13.9$ & $2.62{\pm}0.35$ & $70.2$\\
& 69 & $281.5{\pm}31.8$ & $3.31{\pm}0.37$ & $71.4$\\
\hline
\multirow{2}{2cm}{STG-LDCPPO}
& 33 & $86.5{\pm}11.7$ & $2.16{\pm}0.29$ & $76.8$\\
& 69 & $254.3{\pm}27.1$ & $2.99{\pm}0.32$ & $77.6$\\
\hline
\multirow{2}{2cm}{PI-GR-STG-LDCPPO}
& 33 & $79.1{\pm}10.8$ & $1.98{\pm}0.27$ & $79.2$\\
& 69 & $253.0{\pm}26.8$ & $2.97{\pm}0.31$ & $79.2$\\
\hline
\multirow{2}{2cm}{PISTG-CPO}
& 33 & $89.8{\pm}12.5$ & $2.25{\pm}0.32$ & $73.3$\\
& 69 & $233.9{\pm}26.5$ & $2.75{\pm}0.31$ & $74.2$\\
\hline
\multirow{2}{2cm}{\textbf{PISTG-}\textbf{LDCPPO}}
& 33 & $\mathbf{68.2{\pm}9.4}$ & $\mathbf{1.71{\pm}0.24}$ & $\mathbf{83.1}$\\
& 69 & $\mathbf{240.8{\pm}24.6}$ & $\mathbf{2.83{\pm}0.29}$ & $\mathbf{82.6}$\\

\hline
\end{tabular}
\end{table}

Table~\ref{tab:pv_curtailment} shows that PISTG-LDCPPO consistently maintains low PV curtailment across both test systems while achieving the lowest voltage- and line-related violations.
Although a few baselines yield slightly lower curtailment in individual cases, they exhibit substantially higher network-constraint violations. These results indicate that the proposed policy achieves a better overall balance between local PV utilization and secure network operation.

\subsubsection{Ablation of PISTG and Constraint-Specific Critics}

To isolate the contributions of the two main methodological mechanisms, we conduct ablation studies on the PISTG encoder and the constraint-specific critic design.
Specifically, STG-LDCPPO removes all physics-informed components; PISTG-LDCPPO w/o PI graph replaces the PI adjacency $\tilde{A}$ with the binary topology adjacency $A$; and PISTG-LDCPPO w/o KCL sets $\beta_1=0$ and PISTG-LDCPPO w/o VDrop sets $\beta_2=0$ in Eq.~\eqref{eq:update PISTG}. The reconstruction loss is retained in all PISTG variants.

\begin{table}[!t]
\centering
\scriptsize
\caption{Ablation of PISTG encoder pre-training on the IEEE 33-bus test windows.}
\label{tab:pistg_encoder_ablation}
\renewcommand\arraystretch{1.0}
\setlength{\tabcolsep}{1.0mm}
\begin{tabular}{lcccc}
\hline
Encoder variant
& Volt. MAE
& Flow
& Norm. KCL
& Norm. V-drop \\
& $\times 10^{-3}$ p.u.
& NRMSE \%
& residual $\times 10^{-2}$
& residual $\times 10^{-3}$ \\
\hline
STG encoder
& 7.41
& 6.82
& 4.86
& 7.53 \\
PISTG w/o PI graph
& 5.12
& 5.41
& 2.67
& 4.96 \\
PISTG w/o KCL
& 4.62
& 4.83
& 3.78
& 3.81 \\
PISTG w/o VDrop
& 5.89
& 5.16
& 2.18
& 6.45 \\
Full PISTG
& \textbf{3.68}
& \textbf{4.12}
& \textbf{1.54}
& \textbf{2.91} \\
\hline
\end{tabular}
\end{table}

Table~\ref{tab:pistg_encoder_ablation} shows that the full PISTG encoder achieves the best reconstruction accuracy and physics consistency, while removing the PI graph, KCL loss, or voltage-drop loss degrades the corresponding spatial or physical constraints. Table~\ref{tab:pistg_rl_ablation} further confirms that these encoder improvements translate into better downstream control: compared with STG-LDCPPO, PISTG-LDCPPO reduces the voltage-violation degree and count by $52.8\%$ and $41.3\%$, respectively. Overall, the gains arise from the complementary effects of physics-informed graph propagation and constraint-specific physical regularization.

\begin{table}[!t]
\centering
\scriptsize
\caption{Impact of individual physics-informed ingredients on downstream LDCPPO testing performance.}
\label{tab:pistg_rl_ablation}
\renewcommand\arraystretch{1.0}
\setlength{\tabcolsep}{1.0mm}
\begin{tabular}{|p{1.5cm}|p{1.2cm}|p{1.8cm}|p{1.8cm}|p{1.5cm}|}
\hline
Method
& Operating cost
& Volt. vio. degree
& Volt. vio. count
& Line-overload degree \\
& AUD/day
& p.u./$(5$min$\cdot$bus$)$
& times/$(5$min$\cdot$bus$)$
& \%/$(5$min$\cdot$line$)$ \\
\hline
STG-LDCPPO
& $1414.8 \pm 36.1$
& $(0.53 \pm 0.09)\times 10^{-3}$
& $(0.63 \pm 0.06)\times 10^{-2}$
& $(1.15 \pm 0.14)\times 10^{-3}$ \\ \hline
w/o PI graph
& $1396.7 \pm 37.2$
& $(0.34 \pm 0.05)\times 10^{-3}$
& $(0.46 \pm 0.05)\times 10^{-2}$
& $(1.14 \pm 0.13)\times 10^{-3}$ \\ \hline
w/o KCL
& $1401.2 \pm 38.5$
& $(0.38 \pm 0.06)\times 10^{-3}$
& $(0.50 \pm 0.05)\times 10^{-2}$
& $(1.14 \pm 0.14)\times 10^{-3}$ \\ \hline
w/o VDrop
& $1405.8 \pm 37.9$
& $(0.42 \pm 0.07)\times 10^{-3}$
& $(0.55 \pm 0.06)\times 10^{-2}$
& $(1.15 \pm 0.15)\times 10^{-3}$ \\ \hline
Full
& $\textbf{1382.4} \pm \textbf{38.9}$
& $\textbf{(0.25} \pm \textbf{0.04)}\times 10^{-3}$
& $\textbf{(0.37} \pm \textbf{0.05)}\times 10^{-2}$
& $\textbf{(1.13} \pm \textbf{0.14)}\times 10^{-3}$ \\
\hline
\end{tabular}
\end{table}

\begin{figure}[!t]
\centering

\subfigure[IEEE 33-bus system.]{
    \includegraphics[width=1.0\linewidth]{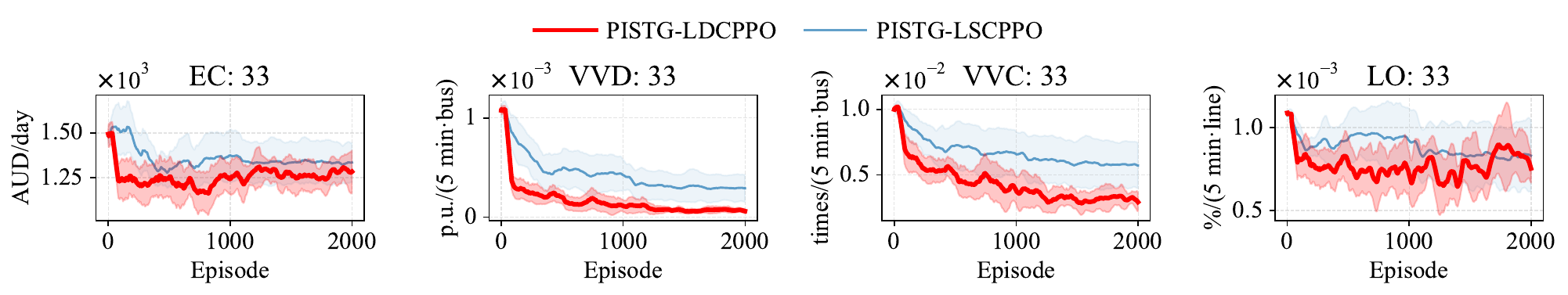}
    \label{fig:R3_comparison_33}
}

\subfigure[IEEE 69-bus system.]{
    \includegraphics[width=1.0\linewidth]{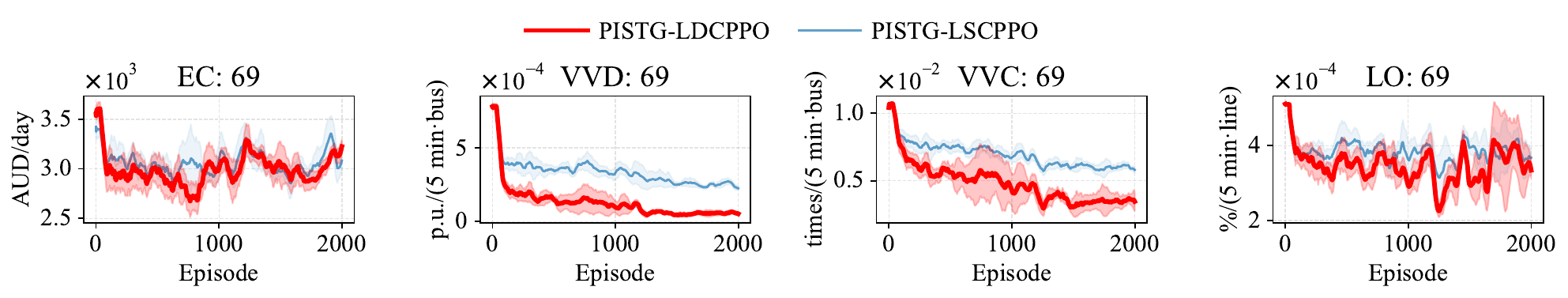}
    \label{fig:R3_comparison_69}
}

\caption{Training-curve comparison between PISTG-LSCPPO and PISTG-LDCPPO.}
\label{fig:R3_comparison}
\end{figure}

We further isolate the contribution of the constraint-specific critic design by comparing PISTG-LDCPPO with PISTG-LSCPPO (Lagrangian Single-Critic PPO), which uses the same PISTG encoder, actor, Lagrangian update, and training/testing protocol but combines the three constraint costs into a single cost critic.
Figures~\ref{fig:R3_comparison_33} and~\ref{fig:R3_comparison_69}, together with Table~\ref{tab:lscppo_comparison}, show that PISTG-LDCPPO consistently reduces empirical constraint violations, especially voltage-violation degree and count, on both systems while maintaining comparable operating costs.
This controlled comparison demonstrates the benefit of estimating heterogeneous constraint returns separately rather than through a single aggregated cost-value estimate.

\begin{table}[!t]
\centering
\scriptsize
\caption{Comparison between PISTG-LSCPPO and PISTG-LDCPPO over 100 test episodes for 5 runs.}
\label{tab:lscppo_comparison}
\renewcommand\arraystretch{1.0}
\setlength{\tabcolsep}{1.0mm}
\begin{tabular}{|p{0.65cm}|p{1.35cm}|p{1.1cm}|p{1.55cm}|p{1.55cm}|p{1.45cm}|}
\hline
System
& Method
& Operating cost
& Volt. vio. degree
& Volt. vio. count
& Line-overload degree \\
&
& AUD/day
& p.u./$(5$min$\cdot$bus$)$
& times/$(5$min$\cdot$bus$)$
& \%/$(5$min$\cdot$line$)$ \\
\hline

33-bus
& PISTG-LSCPPO
& $1408.6 \pm 41.5$
& $(0.52 \pm 0.08)\times 10^{-3}$
& $(0.58 \pm 0.07)\times 10^{-2}$
& $(1.20 \pm 0.15)\times 10^{-3}$ \\
\hline

33-bus
& \textbf{PISTG-LDCPPO}
& $\textbf{1382.4} \pm \textbf{38.9}$
& $\textbf{(0.25} \pm \textbf{0.04)}\times 10^{-3}$
& $\textbf{(0.37} \pm \textbf{0.05)}\times 10^{-2}$
& $\textbf{(1.13} \pm \textbf{0.14)}\times 10^{-3}$ \\
\hline

69-bus
& PISTG-LSCPPO
& $3118.4 \pm 148.7$
& $(0.88 \pm 0.14)\times 10^{-3}$
& $(0.79 \pm 0.12)\times 10^{-2}$
& $(1.28 \pm 0.16)\times 10^{-3}$ \\
\hline

69-bus
& \textbf{PISTG-LDCPPO}
& $\textbf{3060.7} \pm \textbf{135.2}$
& $\textbf{(0.52} \pm \textbf{0.09)}\times 10^{-3}$
& $\textbf{(0.60} \pm \textbf{0.10)}\times 10^{-2}$
& $\textbf{(1.15} \pm \textbf{0.14)}\times 10^{-3}$ \\
\hline
\end{tabular}
\end{table}

\subsubsection{Sensitivity Analysis of PIAM}

To further validate the physics-informed adjacency matrix (PIAM), we conduct two sensitivity studies. The first study evaluates different edge-weight constructions, and the second study examines the effect of the capacity-weight coefficient $\nu$ in PIAM. 

\begin{figure}[!t]
\centering
\includegraphics[width=0.8\linewidth]{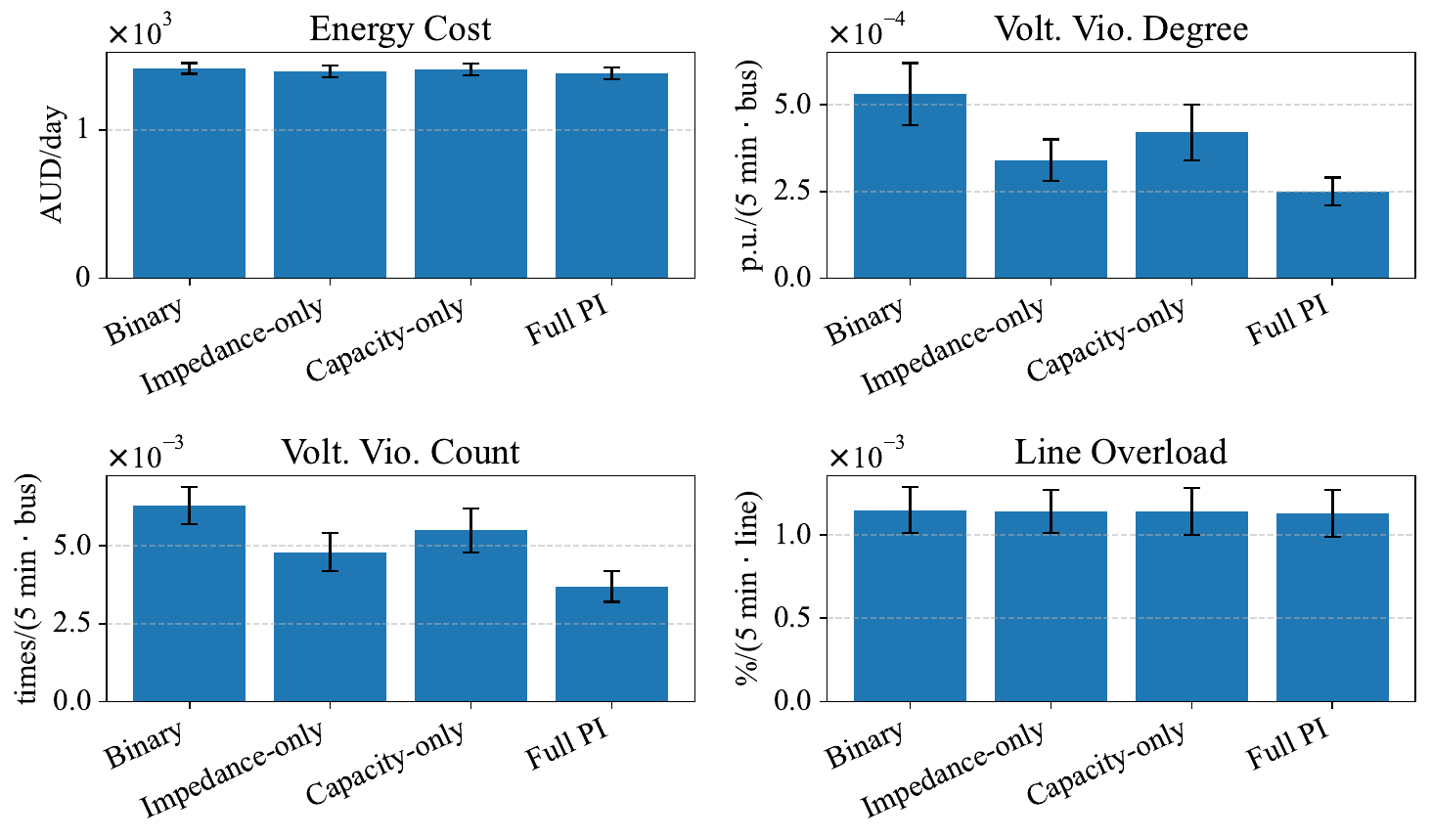}
\caption{Sensitivity analysis of different PIAM edge-weight constructions.}
\label{fig:piam_construction_sensitivity}
\end{figure}

Figure~\ref{fig:piam_construction_sensitivity} compares four adjacency designs: binary topology, impedance-only PI, capacity-only PI, and full PI adjacency. The full PI adjacency achieves the best overall performance, while the two single-factor variants also outperform the binary topology baseline. This confirms that jointly using line impedance and capacity provides a more informative physical graph prior than topology-only or single-factor weighting.

\begin{figure}[!t]
\centering
\includegraphics[width=0.8\linewidth]{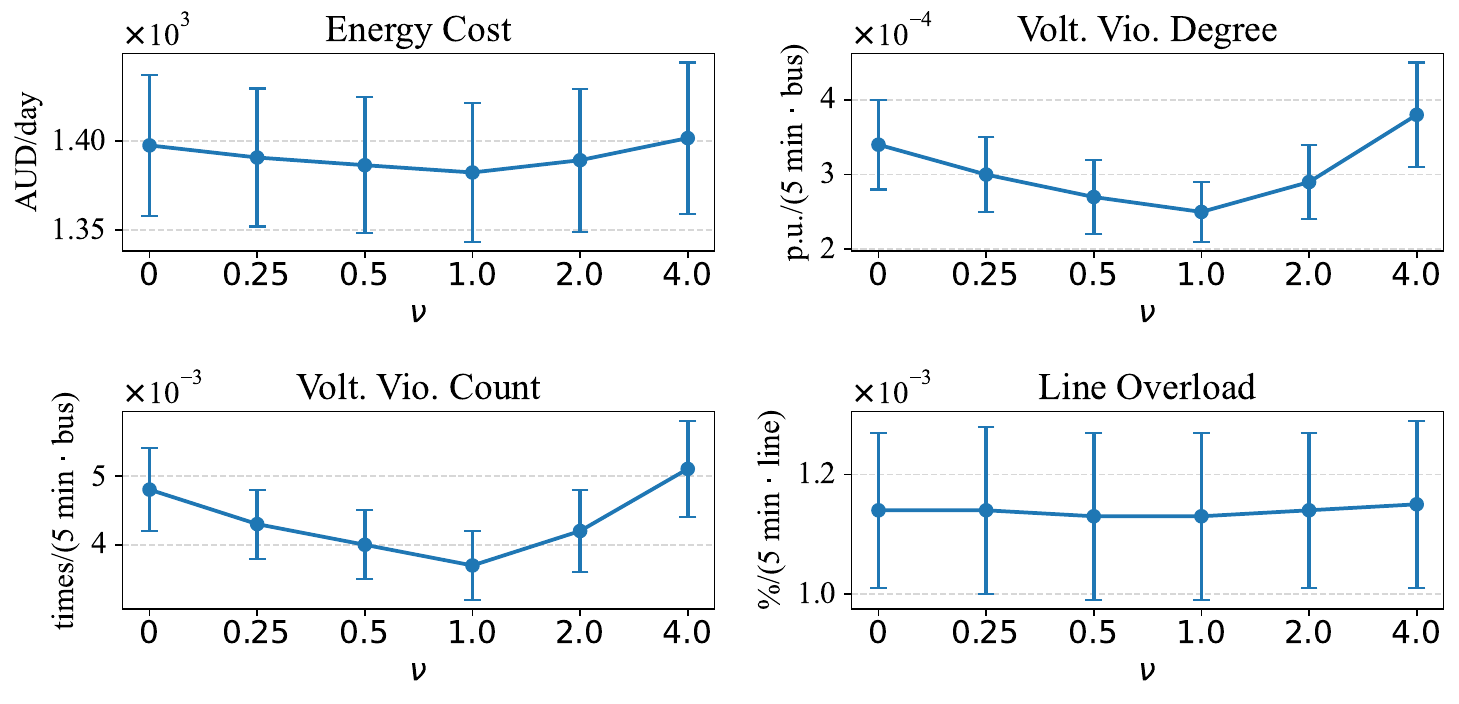}
\caption{Sensitivity analysis of the PIAM capacity-weight coefficient $\nu$.}
\label{fig:piam_nu_sensitivity}
\end{figure}

Figure~\ref{fig:piam_nu_sensitivity} shows that PIAM is robust to the capacity-weight coefficient $\nu$. The performance varies from $0.5$ to $2.0$, with the best overall result at $\nu=1.0$, indicating that PIAM is benefiting from a balanced use of impedance and capacity information.

\begin{table}[!t]
\centering
\scriptsize
\caption{Physical correlation analysis of learned representations.}
\label{tab:physical_correlation}
\renewcommand{\arraystretch}{1.0}
\setlength{\tabcolsep}{1.6mm}
\begin{tabular}{lcccccc}
\hline
\multirow{2}{*}{Encoder}
& \multicolumn{2}{c}{Volt. vio. degree}
& \multicolumn{2}{c}{Volt. vio. count}
& \multicolumn{2}{c}{Line-overload degree} \\
\cline{2-7}
& $\rho$ & $R^2$
& $\rho$ & $R^2$
& $\rho$ & $R^2$ \\
\hline
STG encoder
& 0.71 & 0.49
& 0.67 & 0.44
& 0.54 & 0.29 \\
\textbf{PISTG encoder}
& \textbf{0.89} & \textbf{0.78}
& \textbf{0.84} & \textbf{0.69}
& \textbf{0.58} & \textbf{0.32} \\
\hline
\end{tabular}
\end{table}

\subsubsection{Physical Correlation Analysis of Learned Embeddings}

To further verify whether the proposed PISTG encoder learns physically meaningful representations, we conduct a physical correlation analysis on the learned latent space. Let $z_t$ denote the latent vector generated by an encoder before the policy head. We examine whether $z_t$ preserves information related to three physical stress indicators:
\begin{align}
\chi_t^{\mathrm{VD}} &= \sum_{i\in\mathcal{N}}\left[(V_{i,t}-\overline{V})^+ + (\underline{V}-V_{i,t})^+\right], \\
\chi_t^{\mathrm{VN}} &= \sum_{i\in\mathcal{N}}\mathbb{I}\left[(V_{i,t}>\overline{V})\vee(V_{i,t}<\underline{V})\right], \\
\chi_t^{\mathrm{LO}} &= \sum_{(i,j)\in\mathcal{L}}\left[\sqrt{p_{ij,t}^2+q_{ij,t}^2}-S_{ij}\right]^+ .
\end{align}
For each indicator $k\in\{\mathrm{VD},\mathrm{VN},\mathrm{LO}\}$, we fit a simple linear probe $\widehat{\chi}_t^k=w_k^\top z_t+b_k$ and report the Pearson correlation coefficient $\rho$ and coefficient of determination $R^2$ on testing trajectories. The probe is only used for interpretation and is not involved in policy training or control execution.

As shown in Table~\ref{tab:physical_correlation}, the PISTG embedding is more strongly correlated with voltage-related physical stress than the ordinary STG embedding. This is consistent with Table~\ref{tab:test_results}, where PISTG-LDCPPO achieves lower voltage violations than STG-LDCPPO. The smaller improvement in line-overload degree also matches the weaker line-overload correlation. These results suggest that the PISTG latent space is better aligned with power-flow-related stress indicators, especially voltage stress.

\section{Conclusion and Future Work}\label{Sec:C}
This paper investigates the coordinated operation of CBESSs in a DN with high DER penetration, aiming to improve empirical network-constraint compliance and economic efficiency. 
Our study targets three key challenges: mixed constraint feedback under multi-CBESS constraint-aware learning, insufficient physical awareness in purely data-driven control policies, and nonlocal spatiotemporal interactions under large-scale CBESS dispatch. To address the first challenge, we develop a constraint-specific value-decomposition mechanism that preserves separate long-term learning signals for heterogeneous operational constraints. To address the latter two, we develop a dual-level physics-informed spatiotemporal representation that embeds distribution-network physics into both graph propagation and latent-state learning, enabling the policy to capture physically meaningful nonlocal spatiotemporal dependencies.
Extensive experiments demonstrate that the proposed method substantially reduces empirical voltage-limit and line-loading violations while maintaining competitive operating cost compared with representative DRL and constrained-RL baselines under scenarios with high DER penetration.

Future work will extend the proposed framework toward more practical deployment scenarios, including three-phase unbalanced and reconfigurable distribution networks, cooperative multi-agent CBESS coordination under local observations and limited communication, and operation at shorter decision intervals with higher-resolution measurements, and under noisy, delayed, or incomplete measurements. We will further develop uncertainty-aware models and an execution-time safety-assurance layer that validates learned actions through online power-flow analysis and, when necessary, applies corrective measures such as inverter reactive-power support, conservative CBESS redispatch, or limited PV curtailment.

\bibliography{reference}
\bibliographystyle{IEEEtran}

\begin{IEEEbiography}
[{\includegraphics[width=1in,height=1.25in,clip,keepaspectratio]{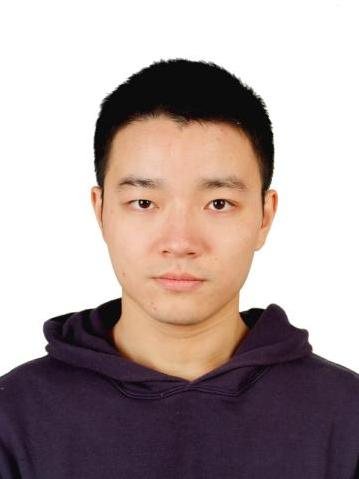}}] 
{Chenghao Huang}
received the B.E. degree in Software Engineering and the M.S. degree in Computer Science and Engineering from the University of Electronic Science and Technology of China, Chengdu, China, in 2020 and 2023, respectively. He is currently pursuing the Ph.D. degree in the Faculty of Information Technology, Monash University, Melbourne, VIC, Australia. His research interests deep learning, federated learning, reinforcement learning and smart grids.
\end{IEEEbiography}

\begin{IEEEbiography}
[{\includegraphics[width=1in,height=1.25in,clip,keepaspectratio]{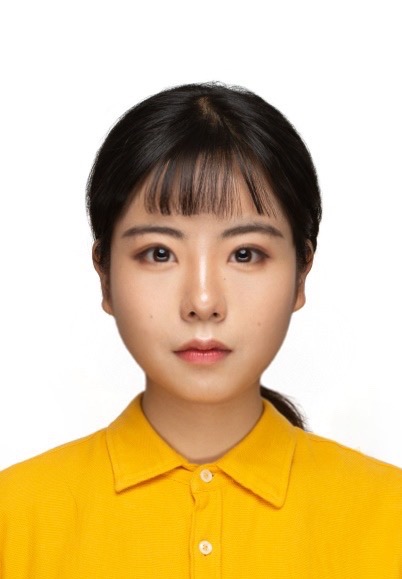}}] 
{Yue Yang} received the B.A. in Mathematics and Computer Science from Mount Holyoke College, Massachusetts, USA, in 2017. She received her Ph.D. in Information Technology from Monash University, Melbourne, VIC, Australia, in 2026. Her research interests include include reinforcement learning, large language model training and inference, with their applications.
\end{IEEEbiography}

\begin{IEEEbiography}
[{\includegraphics[width=1in,height=1.25in,clip,keepaspectratio]{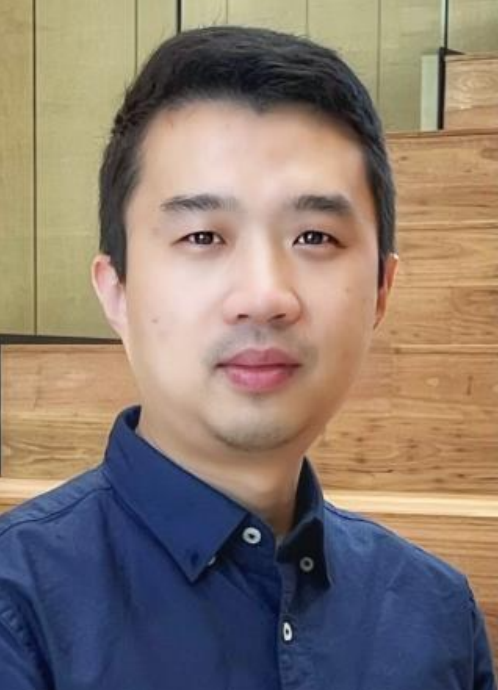}}]{Hao Wang} (M'16) received his Ph.D. in Information Engineering from The Chinese University of Hong Kong, Hong Kong, in 2016. He was a Postdoctoral Research Fellow at Stanford University, Stanford, CA, USA, and a Washington Research Foundation Innovation Fellow at the University of Washington, Seattle, WA, USA. 

He is currently a Senior Lecturer in the Department of Data Science and AI, Faculty of IT, Monash University, Melbourne, VIC, Australia. He received Washington Research Foundation Innovation Fellowship 2016-2018, ARC Discovery Early Career Researcher Award (DECRA) 2023-2025, First Prize Paper Award from IEEE Industry Applications Magazine 2024, Best Vision Paper Award from ACM SIGSPATIAL 2024, Best Conference Paper from IEEE PESGM 2024, Best Paper Award at IEEE SmartGridComm 2025, Best Special Theme Paper at EMNLP 2025, Best Paper Award at IEEE EPEC 2025, IJPEM-GT Award from International Journal of Precision Engineering and Manufacturing-Green Technology in 2025. His research interests include optimization, machine learning, and data analytics for power and energy systems.
\end{IEEEbiography}

\vfill
\end{document}